\documentclass{article}

\usepackage{natbib}
\usepackage{amsmath}
\usepackage{amssymb}
\usepackage{graphicx}
\usepackage{subfigure}
\graphicspath{{./images/},{./figs/},{./plots/}}

\usepackage{authblk}


\begin{document}

\setcounter{page}{1}










\title{Learning, Generalization, and Functional Entropy in Random
  Automata Networks}
\author[1]{Alireza Goudarzi}
\author[2]{Christof Teuscher}
\author[3]{Natali Gulbahce}
\author[4,5]{Thimo Rohlf}
\affil[1]{Department of Computer Science,\\ Portland State University\\
1900 SW $4^{th}$ Ave, Portland, OR 97201, USA\\ E-mail: alirezag@cs.pdx.edu}
\affil[2]{Department of Electrical and Computer Engineering,\\
Portland State University\\
1900 SW $4^{th}$ Ave, Portland, OR 97201, USA}
\affil[3]{Department of Cellular and Molecular Pharmacology\\ University of California, San Francisco (UCSF), 1700 $4^{th}$, San~Francisco, CA 94158, USA}
\affil[4]{Interdisciplinary Center for Bioinformatics, University Leipzig, Haertelstr. 16-18, D-04107 Leipzig, Germany}
\affil[5]{
Epigenomics Project, iSSB, Genopole Campus 1, 5 Rue Henri Desbrueres, F-91034 Evry, France\\
Max-Planck-Institute for Mathematics in the Sciences, Inselstr. 22, D-04103 Leipzig, Germany}

\maketitle

\begin{abstract}

  It has been shown
  \citep{broeck90:physicalreview,patarnello87:europhys} that
  feedforward Boolean networks can learn to perform specific simple
  tasks and generalize well if only a subset of the learning examples
  is provided for learning. Here, we extend this body of work and show
  experimentally that random Boolean networks (RBNs), where both the
  interconnections and the Boolean transfer functions are chosen at
  random initially, can be evolved by using a state-topology evolution
  to solve simple tasks.  We measure the learning and generalization
  performance, investigate the influence of the average node
  connectivity $K$, the system size $N$, and introduce a new measure
  that allows to better describe the network's learning and
  generalization behavior. We show that the connectivity of the
  maximum entropy networks scales as a power-law of the system size
  $N$. Our results show that networks with higher average connectivity
  $K$ (supercritical) achieve higher memorization and partial
  generalization. However, near critical connectivity, the networks
  show a higher perfect generalization on the even-odd task.

\end{abstract}

\section{Introduction}
\label{sec:intro}
Pattern recognition is a task primates are generally very good at
while machines are not so much.  Examples are the recognition of human
faces or the recognition of handwritten characters. The scientific
disciplines of machine learning and computational learning theory have
taken on the challenge of pattern recognition since the early days of
modern computer science. A wide variety of very sophisticated and
powerful algorithms and tools currently exist \citep{bishop06}. In
this paper we are going back to some of the roots and address the
challenge of learning with networks of simple Boolean logic gates. To
the best of our knowledge, Alan Turing was the first person to explore
the possibility of learning with simple NAND gates in his long
forgotten 1948 paper, which was published much later
\citep{turing48,teuscher01:conn}. One of the earliest attempts to
classify patterns by machine came from \cite{selfridge:58}, and
\cite{neisser60:sciam}. Later, many have explored random logical nets
made up from Boolean or threshold (McCulloch-Pitts) neurons:
\citep{rozonoer69,amari71,aleksander1984,aleksander98,aleksander1973}.
\cite{martland87} showed that it is possible to predict the activity
of a boolean network with randomly connected inputs, if the
characteristics of the boolean neurons can be described
probabilistically. In a second paper, \cite{martland87b} illustrated
how the boolean networks are used to store and retrieve patterns and
even pattern sequences auto-associatively. Seminal contributions on
random Boolean networks came from
\cite{kauffman69,kauffman93,kauffman84} and Weisbuch
\cite{weisbuch89,weisbuch91}.

In 1987, Carnevali and Patarnello
\citep{patarnello87:europhys,carnevali87:europhys}, used {\em
  simulated annealing} and in 1989 also {\em genetic algorithms}
\citep{patarnello89b:aleksander} as a global stochastic optimization
technique to train feedforward Boolean networks to solve computational
tasks. They showed that such networks can indeed be trained to
recognize and generalize patterns.  \cite{broeck90:physicalreview}
also investigated the learning process in feedforward Boolean networks
and discovered their amazing ability to generalize.

\cite{teuscher07:ddaysrbn} presented
preliminary results that true RBNs, i.e., Boolean networks with
recurrent connections, can also be trained to learn and generalize
computational tasks. They further hypothesized that the performance is
best around the critical connectivity $K = 2$.

In the current paper, we extend and generalize Patarnello and
Carnevali's results to random Boolean networks (RBNs) and use genetic
algorithms to evolve both the network topology and the node transfer
functions to solve a simple task.  Our work is mainly motivated by the
application of RBNs in the context of emerging nanoscale electronics
\citep{teuscher09:ijnmc}. Such networks are particularly appealing for
that application because of their simplicity. However, what is lacking
is a solid approach that allows to train such systems for performing
specific operations. Similar ideas have been explored with none-RBN
building blocks by \cite{tour02} and by \cite{lawson06}. One of the
broader goals we have is to systematically explore the relationship
between generalization and learning (or memorization) as a function of
the system size, the connectivity $K$, the size of the input space,
the size of the training sample, and the type of the problem to be
solved. In the current paper, we restrict ourselves to look at the
influence of the system size $N$ and of connectivity $K$ on the
learning and generalization capabilities. In the case of emerging
electronics, such as for example self-assembled nanowire networks use
to compute simple functions, we are interested to find the smallest
network with the lowest connectivity that can learn how to solve the
task with the least number of patterns presented.

\section{Random Boolean Networks}
\label{sec:rbn}
A {\sl random Boolean network} (RBN)
\cite{kauffman69,kauffman84,kauffman93} is a discrete dynamical system
composed of $N$ nodes, also called {\sl automata}, {\sl elements} or
{\sl cells}. Each automaton is a Boolean variable with two possible
states: $\{0,1\}$, and the dynamics is such that

\begin{equation}
{\bf F}:\{0,1\}^N\mapsto \{0,1\}^N, 
\label{globalmap}
\end{equation} 

where ${\bf F}=(f_1,...,f_i,...,f_N)$, and each $f_i$ is represented
by a look-up table of $K_i$ inputs randomly chosen from the set of $N$
nodes. Initially, $K_i$ neighbors and a look-up table are assigned to
each node at random.  Note that $K_i$ (i.e., the fan-in) can refer to
the {\sl exact} or to the {\sl average} number of incoming connections
per node. In this paper we use $K$ to refer to the average
connectivity.

A node state $ \sigma_i^t \in \{0,1\}$ is updated using its
corresponding Boolean function:

\begin{equation}
\sigma_i^{t+1} = f_i(\sigma_{i_1}^t,\sigma_{i_2}^t, ... ,\sigma_{i_{K_i}}^t).
\label{update}
\end{equation}

These Boolean functions are commonly represented by {\sl look-up
  tables} (LUTs), which associate a $1$-bit output (the node's future
state) to each possible $K$-bit input configuration. The table's
out-column is called the {\sl rule} of the node. Note that even though
the LUTs of a RBN map well on an FPGA or other memory-based
architectures, the random interconnect in general does not.

We randomly initialize the states of the nodes (initial condition of
the RBN). The nodes are updated synchronously using their
corresponding Boolean functions. Other updating schemes exist, see for
example \citep{gershenson2003:alife} for an overview. Synchronous
random Boolean networks as introduced by Kauffman are commonly called
$NK$ {\sl networks} or {\sl models}.

\begin{figure}
  \centering \includegraphics[width=.95\textwidth]{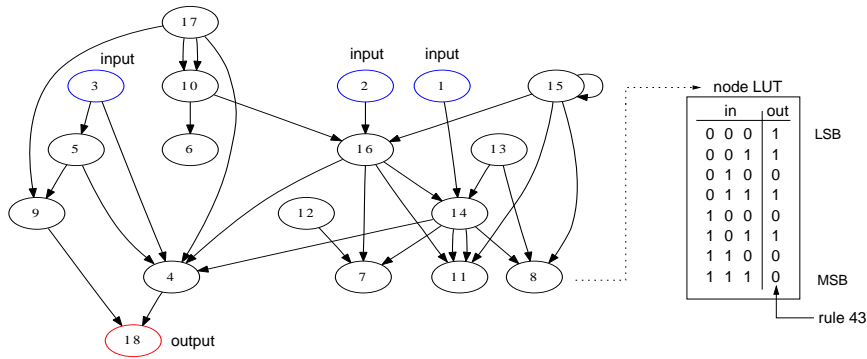}
  \caption{Illustration of an $18$-node RBN with 3 input nodes (node
    IDs 1, 2, and 3, colored in blue) and 1 output node (node ID 18,
    colored in red). The average connectivity is $K=2.5$. The node
    rules are commonly represented by {\sl lookup-tables} (LUTs),
    which associate a $1$-bit output (the node's future state) to each
    possible $K$-bit input configuration. The table's out-column is
    commonly called the {\sl rule} of the node.}
  \label{fig:network_example}
\end{figure} 

The ``classical'' RBN is a closed system without explicit inputs and
outputs. In order to solve tasks that involve inputs and outputs, we
modify the classical model and add $I$ input nodes and designate $O$
nodes as output nodes. The input nodes have no logical function and
simply serve to distribute the input signals to any number of
{randomly chosen} nodes in the network.  On the other
hand, the output nodes are just like any other network node, i.e.,
with a Boolean transfer function, except that their state can be read
from outside the network. {The network is constructed
  in a random unbiased process in which we pick $L=N\times K$ pairs of
  source and destination nodes from the network and connect them with
  probability $p=0.5$. This construction results in a binomial
  in-degree distribution in the initial network population \citep{Erdos:1959p1849}. The source
  nodes can be any of the input nodes, compute nodes, or the output
  nodes and the destination nodes can be chosen only from the compute
  nodes and the output nodes.} Figure \ref{fig:network_example} shows
an $18$-node RBN with $3$ input nodes and $1$ output node.

\section{Functional Entropy}
\label{sec:functionalentropy}
 Any given network in the space of all possible networks processes
 information and realizes a particular function. Naturally, the task
 of GAs (or any other search technique) is to only search in the space
 of possible networks and to find networks that realize a desired
 function, such as for example the even-odd task. Therefore, the
 learning capability, with respect to the entire class of functions,
 can be interpreted as the frequency of the realization of all
 possible functions. In our case, that means the class of Boolean
 functions with three inputs by using a class of ``computers,'' i.e.,
 the Boolean networks. \cite{broeck90:physicalreview} and
 \cite{Amirikian:1994p42} investigated the {\em phase volume} of a
 function, which they defined as the number of networks that realize a
 given function. Thus, the entropy of the functions realized by all
 possible networks is an indicator of the richness of the
 computational power of the networks. We extend this concept to the
 class of random Automata networks $G(N,K)$ characterized using two
 parameters: the size of the network $N$ and the average connectivity
 $K$. We call this the {\sl functional entropy} of the $NK$ landscape.

\begin{figure}
 \subfigure[$N=20$, $I=3$]{
  \includegraphics[width=.495\textwidth]{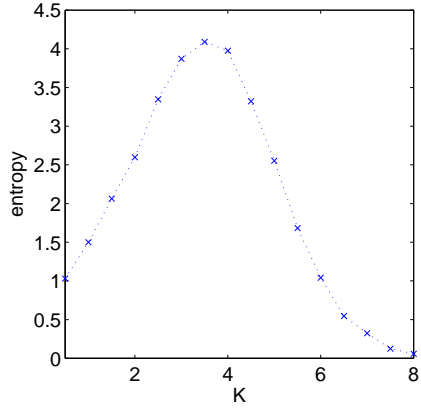}
\label{fig:eni3n20}
}
\subfigure[$N=20$, $I=5$]{
  \includegraphics[width=.495\textwidth]{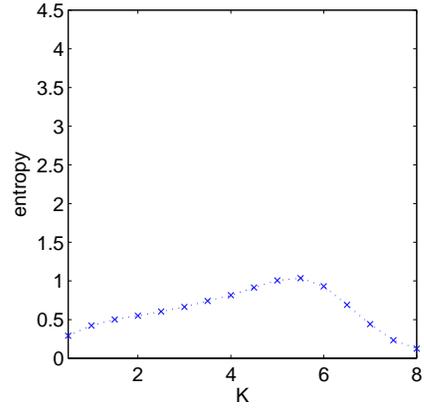}
\label{fig:eni5n20}
}
\subfigure[$N=100$, $I=3$]{
  \includegraphics[width=.495\textwidth]{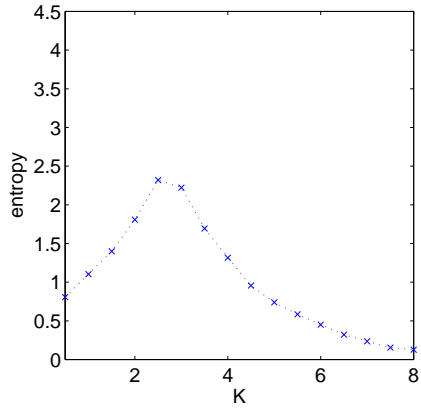}
\label{fig:eni3n100}
}
\subfigure[$N=100$, $I=5$]{
  \includegraphics[width=.495\textwidth]{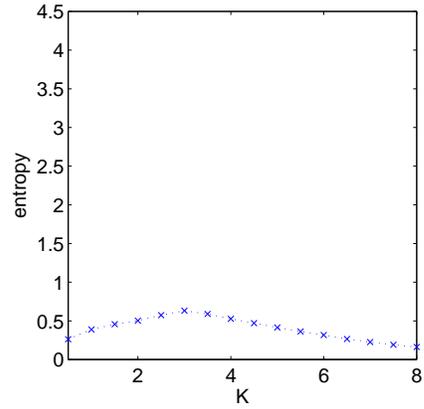}
\label{fig:eni5n100}
}
\caption{Entropy landscape of $N=20$ and $N=100$ networks. The entropy
  is calculated by simulating $10,000$ networks for each $K$. The
  maximum entropy for $I=3$ and $I=5$ is $8$ bits and $32$ bits
  respectively. Due to exponential probability distribution and the
  inadequacy of sampling over the space of networks, the actual values
  are much lower then the theoretical values. However, the position of
  the maximum empirical entropy as a function of $K$ is valid due to
  unbiased sampling of the space.}
  \label{fig:entropyn20n100}
\end{figure}

 Figure~\ref{fig:entropyn20n100} shows the landscape of the functional
 entropy for networks of $N=20$ and $N=100$ with an average
 connectivity of $0.5\le K\le 8$. To calculate the functional entropy,
 we create $10,000$ networks with a given $N$ and $K$. We then
 simulate the networks to determine the function each of the networks
 is computing. The entropy can then be simply calculated using:

\begin{equation}
S_{G(N,K)}=-\sum_{i}{p_ilog_2p_i}.
\end{equation}

Here, $p_i$ is the probability of the function $i$ being realized by
the network of $N$ nodes and $K$ connectivity. For $I=3$, there are
$256$ different Boolean functions. Thus, the maximum entropy of the
space is $8$. This maximum entropy is achievable only if all $256$
functions are realized with equal probability. This is, however, not
the case because the distribution of the functions is not uniform in
general. Also, the space of possible networks cannot be adequately
represented in $10,000$ samples. However, our sampling is good enough
to estimate a comparative richness of the functional entropy of
different classes of networks. For example for $N=20$, the peak of
the entropy in the space of Boolean functions with three inputs lies
at $K=3.5$, whereas for the class of five-input functions, this peak
is at $K=5.5$ (Figures~\ref{fig:eni3n20} and \ref{fig:eni5n20}.) For
$N=100$, the peak of the entropy for three-input and five-input
functions is at $K=2.5$ and $K=3.0$ respectively
(Figure~\ref{fig:eni3n100} and \ref{fig:eni5n100}.) The lower $K$
values of the maximum entropy for larger networks suggests that as $N$
increases, the networks will have their highest capacity in a lower
connectivity range. 

\begin{figure}
\centering
 \subfigure[Maximum entropy scaling on a linear scale.]{
  \includegraphics[width=.69\textwidth]{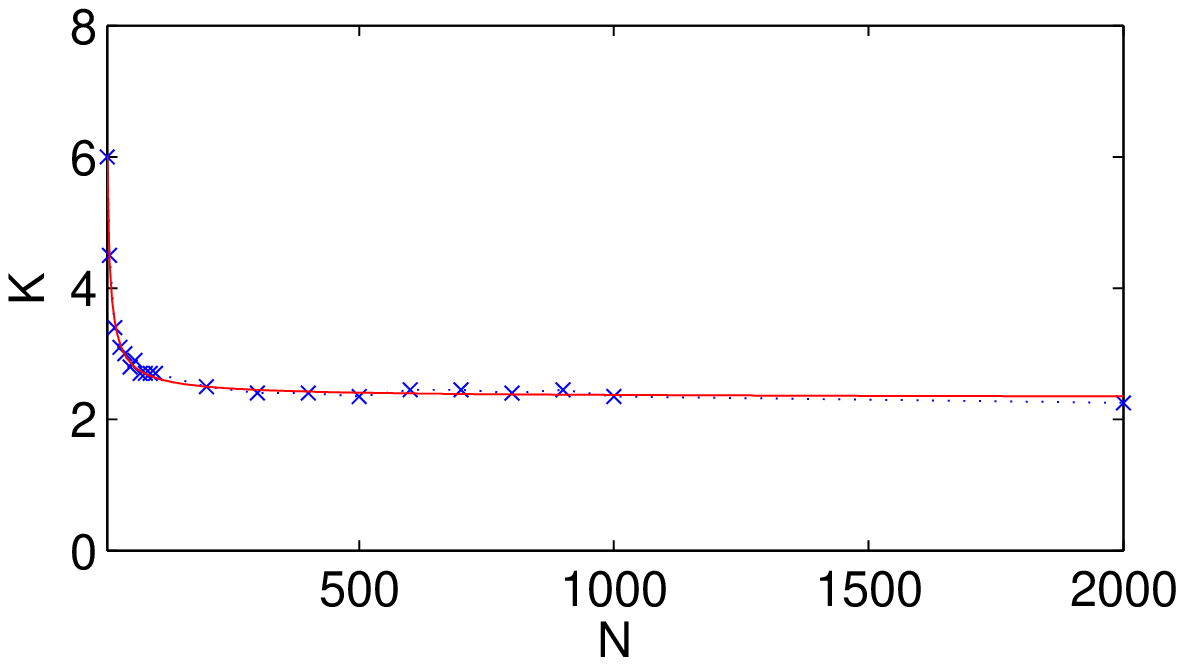}
\label{fig:linear_ijaacs}
}\\
\subfigure[Maximum entropy scaling on a log-log scale.]{
  \includegraphics[width=.69\textwidth]{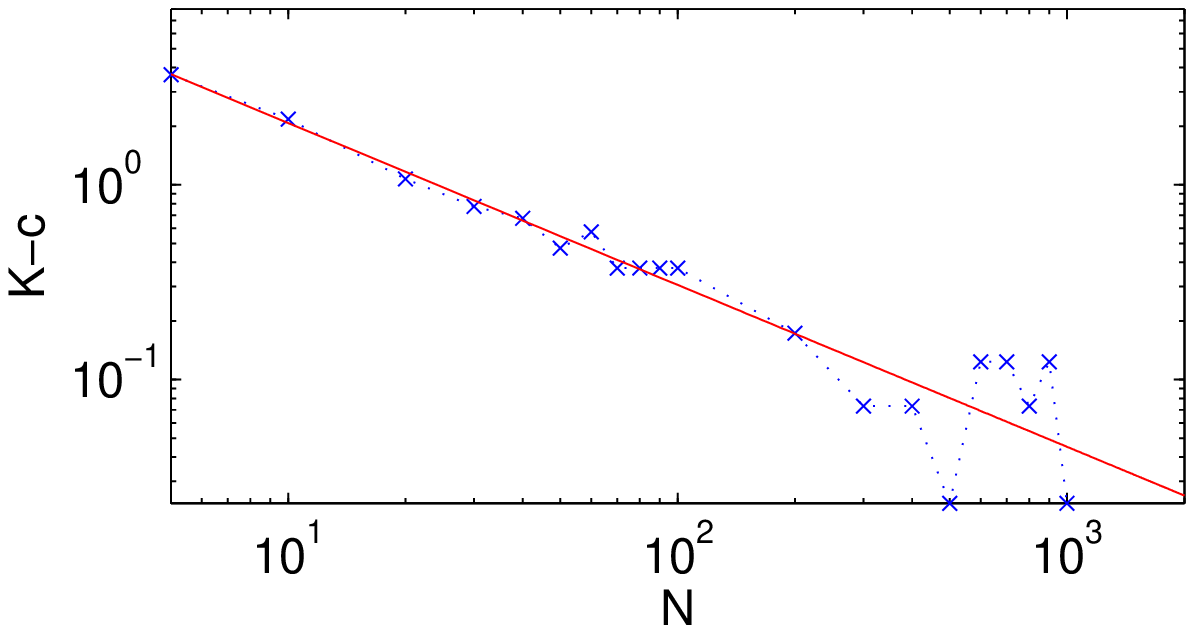}
\label{fig:log-log-2_ijaacs}
}
\caption{\subref{fig:linear_ijaacs} The connectivity of the maximum
  entropy networks scales as a power-law of the system size $N$
  according to
  Equation~\ref{eq:power-law}. \subref{fig:log-log-2_ijaacs} is
  generated by subtracting $c=2.32$ from the data points.}
  \label{fig:maxentropyscaling}
\end{figure}

To study how the maximum attainable functional entropy changes as a
function of $N$, we created networks with sizes $5\le N\le 2000$ and
$0.5\le K\le 8.0$ and determined the maximum of the entropy landscape
as a function of $K$. Figures~\ref{fig:linear_ijaacs} and
\ref{fig:log-log-2_ijaacs} show the scaling of the maximum functional
entropy as a function of $K$ on linear and log-log scales
respectively. As one can see, the data points from the simulations
follow a power-law of the form:
\begin{equation}
K=aN^b+c,
\label{eq:power-law}
\end{equation}
where, $a=14.06$, $b=-0.83$, and $c=2.32$. The solid line in the plots
shows the fitted power-law equation. In
Figure~\ref{fig:log-log-2_ijaacs}, the straight line is the result of
subtracting $c=2.32$ from the equation and from the data points.

{Studying the functional entropy of the network
  ensembles reveals features of the network fitness landscape in the
  context of task solving. In Section~\ref{sec:cummeasures}, we will
  see how functional entropy explains the result of the cumulative
  performance measures.}

\section{Experimental Setup}
\label{sec:expsetup}
We use {\em genetic algorithms} (GAs) to train the RBNs to solve the
even-odd {task}, the mapping task, and the bitwise
AND task. The {\em even-odd task} consists of determining if an
$l-$bit input has an even or an odd number of $1$s in the input. If
the number of $1$s is an odd number, the output of the network must be
$1$, {and} $0$ otherwise. This task is admittedly
rather trivial if one allows for counting the number of $1$s. Also, if
enough links are assigned to a single RBN node, the task can be solved
with a single node since all the combinations can be enumerated in the
look-up table. However, we are not interested to find such trivial
solutions, instead, we look for networks that are able to generalize
well if only a subset of the input patterns is presented during the
training phase. In Section \ref{sec:exp1} we also use the {\em bitwise
  AND task}, which does exactly what its name suggests, i.e., form the
logical AND operation bit by bit with two $l-$bit inputs and one
$l-$bit output. The {\em mapping task} is used in Section
\ref{sec:exp2} and consists of a $l-$bit input and an $l-$bit
output. The output must have the same number of $l-$bits as the input,
but not necessarily in the same order.  Throughout the rest of the
paper, we use $I$ to refer to the total number of input bits to the
network. For example, the bitwise AND for two $3$-bit inputs is a
problem with $I=6$ inputs.

To apply GAs, we encode the network into a bit-stream that consists of
both the network's adjacency matrix and the Boolean transfer functions
for each node. {We represent the adjacency matrix in a
  list of source and destination node IDs of each link. We then append
  this list with the look-up tables for each node's transfer
  function. Note that the index to the beginning and the end of the
  look-up table for each node can be calculated by knowing the node
  index and node in-degree.} The genetic operators consist of a
mutation and a one-point crossover operator that are applied to the
genotypes in the network population. The mutation operator picks a
random location in the genome and performs either of the following two
operations, depending on the content of that location:
\begin{enumerate}
\item If the location points to a source or a destination node of a
  link, we randomly replace it with a pointer to a new node in the
  network.
\item If the location contains a bit in the LUT, we flip that bit. 
\end{enumerate}
We perform crossover by choosing a random location in the two genomes
and then exchange the contents of the two genomes split at that
point. We further define a fitness function $f$ and a generalization
function $g$. For an input space $M'$ of size $m'$ and an input sample
$M$ of size $m$ we write: $E_M = \frac{1}{m}\sum_{j\in M}{d(j)}$ with
$f=1-E_M$, where $d(j)$ is the Hamming distance between the network
output for the $j^{th}$ input in the random sample from the input
space and the expected network output for that input. Similarly, we
write: $E_{M'} = \frac{1}{n}\sum_{j\in M'}{d(j)}$ with $g=1-E_{M'}$,
where $d(i)$ is the Hamming distance between the network output for
the $i^{th}$ input from the entire input space and the expected
network output for that input.

The simple genetic algorithm we use is as following:
\begin{enumerate}
\item Create a random initial population of $S$ networks.
\item Evaluate the performance of the networks on a random sample of
  the input space.
\item Apply the genetic operators to obtain a new population.
\item {For the selection, we use a deterministic
    tournament in which pairs of individuals are selected randomly and
    the better of the two will make it into the offspring population.}
\item Continue with steps 2 and 3 until at least one of the networks
  achieves a perfect fitness or after $G_{max}$ generations are
  reached.
\end{enumerate}

To optimize feedforward networks (see Section \ref{sec:exp1}), we have
to make sure that the mutation and crossover operators do not violate
the feedforward topology of the network. We add an order attribute to
each node on the network and the nodes accept connections only from
lower order nodes.

Since RBNs have recurrent connections, their rich dynamics need to be
taken into account when solving tasks, and in particular interpreting
output signals. Their finite and deterministic behavior guarantees
that a network will fall into a (periodic or fixed point) attractor
after a finite number of steps. The transient length depends on the
network's average connectivity $K$ and the network size $N$
\citep{kauffman93}. For our simulations, we run the networks long
enough until they reach an attractor. Based on \citep{kauffman93}, we
run our networks (with $k<5$) for $2N$ time steps to reach an
attractor. However, due to potentially ambiguous outputs on periodic
attractors, we further calculate the average activation of the output
nodes over a number of time steps equal to the size $N$ of the network
and consider the activity level as $1$ if at least half of the time
the output is $1$, otherwise the activity will be $0$. A similar
technique was used successfully in \citep{teuscher01:conn}.

\section{Training and Network Performance Definitions}
\label{sec:definitions}
\cite{patarnello87:europhys} introduced the
notion of {\em learning probability} as a way of describing the
learning and generalization capability of their feedforward
networks. They defined the learning probability as the probability of
the training process yielding a network with perfect generalization,
given that the training achieves perfect fitness on a sample of the
input space.  

The learning probability is expressed as a function of the fraction of
the input space, $s=\frac{m}{m'}$, used during the training. To
calculate this measure in a robust way, we run the training process
$r$ times and store both the fitness $f$ and the generalization $g$
values. We define the learning probability as a function of $s$,
$\delta(s)= Pr(g=1|f=1)=\frac{\alpha'(s)}{\alpha(s)}$, where
$\alpha(s)=Pr(f=1)$ is the {\textit{perfect training
    likelihood}, i.e., the} probability of achieving a perfect fitness
{($f=1$)} after training, and $\alpha'(s)=Pr(g=1)$ is
the probability of obtaining a perfect fitness in generalization,
{($g=1$)}.  In the following sections, we will define
new measures to evaluate the network performance more effectively.

One can say that the probabilistic measures, such as the learning
probability described above, only focus on the perfect cases and hence
describe the performance of the training process rather than the
effect of the training on the network performance.  Thus, we define
the {\em mean training score} as $\beta(s)=\frac{1}{r}\sum_r
f_{final}$ and the {\em mean generalization score} as
$\beta'(s)=\frac{1}{r}\sum_r g_{final}$, where {$r$
  is the number of evolutionary runs,} and $f_{final}$ and $g_{final}$
are the training fitness and the generalization fitness of the best
networks respectively at the end of training.

To compare the overall network performance for different training
sample sizes, we introduce a {\em cumulative measure} for all four
measures as defined above. The cumulative measure is obtained by a
simple trapezoidal integration \citep{wittaker69} to calculate the area
under the curve for the learning probability, the perfect training
likelihood, the mean generalization score, and mean training score.

\section{Learning in Feedforward Boolean Networks}
\label{sec:exp1}
\begin{figure}
  \centering \includegraphics[width=.495\textwidth]{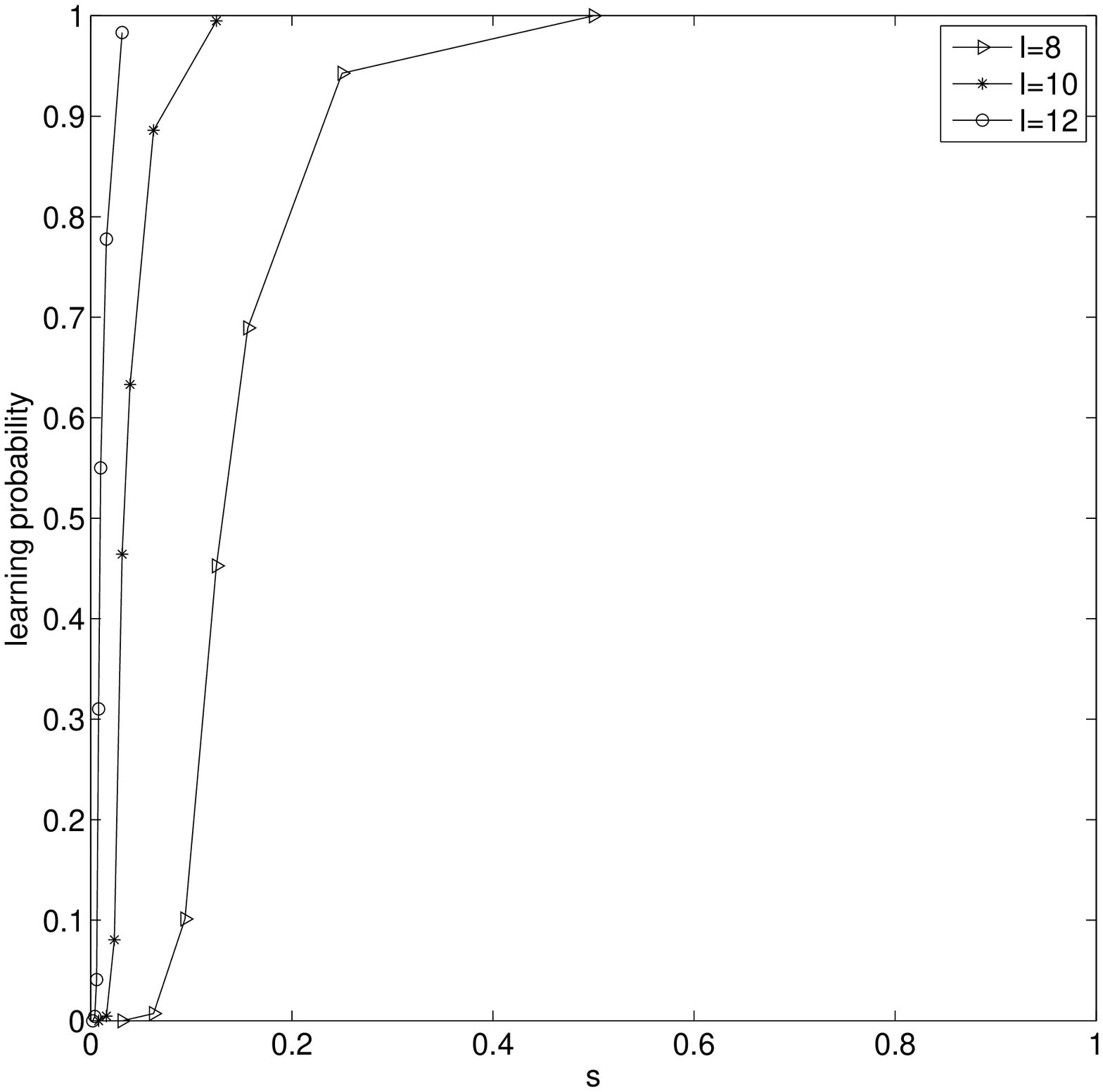}
  \centering \includegraphics[width=.495\textwidth]{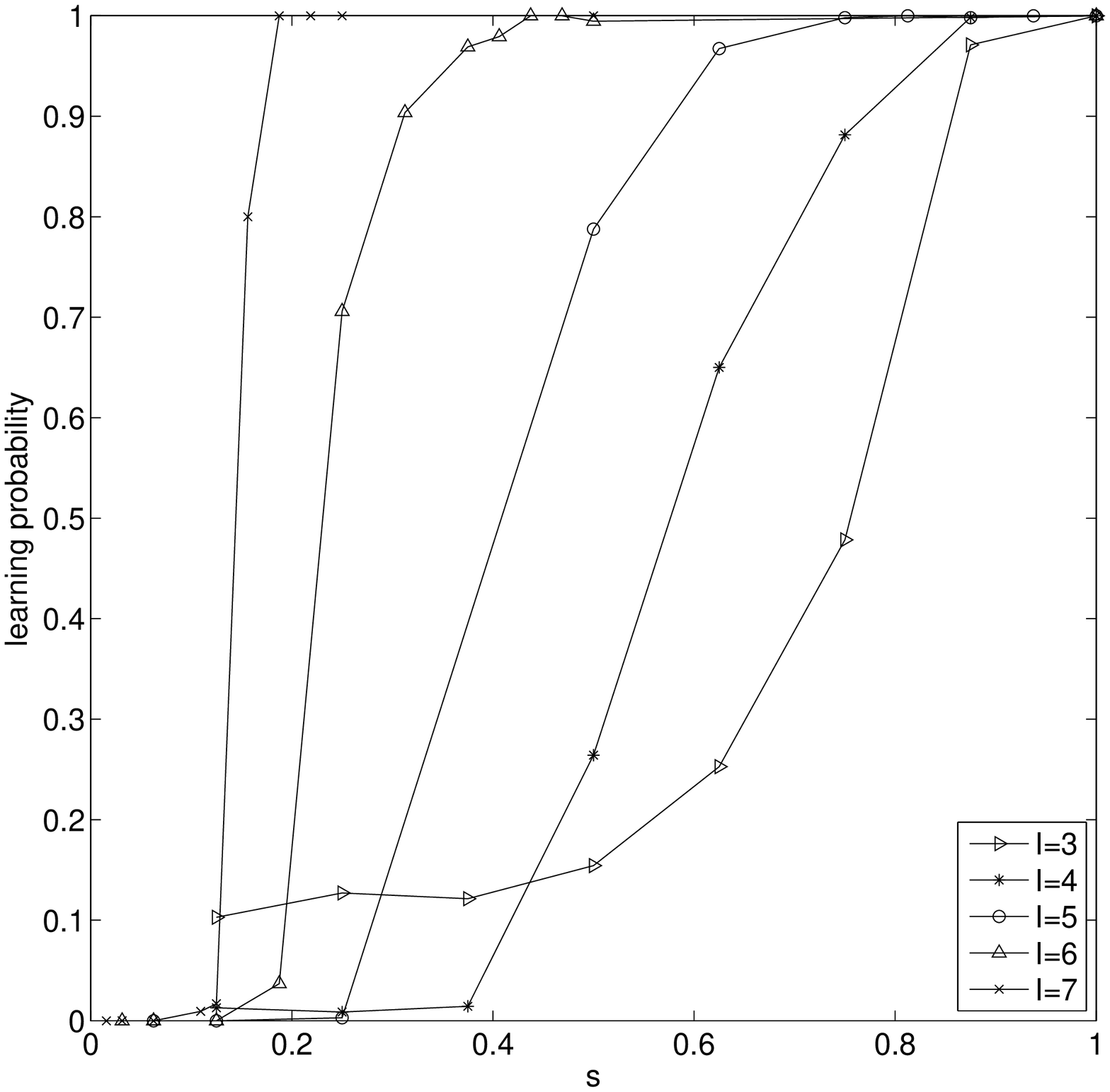}
  \caption{LEFT: The learning probability of feedforward networks on
    the bitwise AND task for different input sizes
    $I$. $s=\frac{m}{m'}$ is the fraction of the input space used in
    training. As $I$ increases, the learning process requires a
    smaller fraction of the input space during the training to achieve
    a perfect learning probability. RIGHT: The learning probability of
    feedforward networks on the even-odd task for various input sizes
    $I$. As $I$ increases, the learning process requires a smaller
    fraction of the input space during the training to achieve a
    perfect learning probability. For $I=3$, some of the networks can
    correctly classify a significant number of patterns without
    training because the task is too easy. For both plots: $N = 50$,
    $K = 2$, $G_{max} = 3000$, initial population size $= 50$,
    crossover rate $= 0.6$, mutation rate $= 0.3$. The GA was repeated
    over $700$ runs.}
  \label{fig:fig7and8}
\end{figure}

The goal of this first experiment was to simply replicate the results that was reported in \citep{patarnello89b:aleksander} with
feedforward Boolean networks. Figure \ref{fig:fig7and8} shows the
learning probability of such networks on the even-odd (RIGHT) and the
bitwise AND task (LEFT) for $K = 2$ networks. We observe that as the
size $I$ of the input space increases, the training process requires a
smaller number of training examples to achieve a perfect learning
probability. For $I=3$, some of the networks can solve a significant
number of patterns without training because the task is too easy. We
have initially determined the GA parameters (see figure legends), such
as the mutation rate and the maximum number of generations
experimentally, depending on how quickly we achieved perfect fitness
on average. We have found the GA to be very robust against parameter
variations for our tasks. These result shown in Figure
\ref{fig:fig7and8} directly confirm Patarnello and Carnevali's
\citep{patarnello89b:aleksander} experiments.

\section{Learning in RBNs}
\label{sec:exp2}
\begin{figure}
  \includegraphics[width=.495\textwidth]{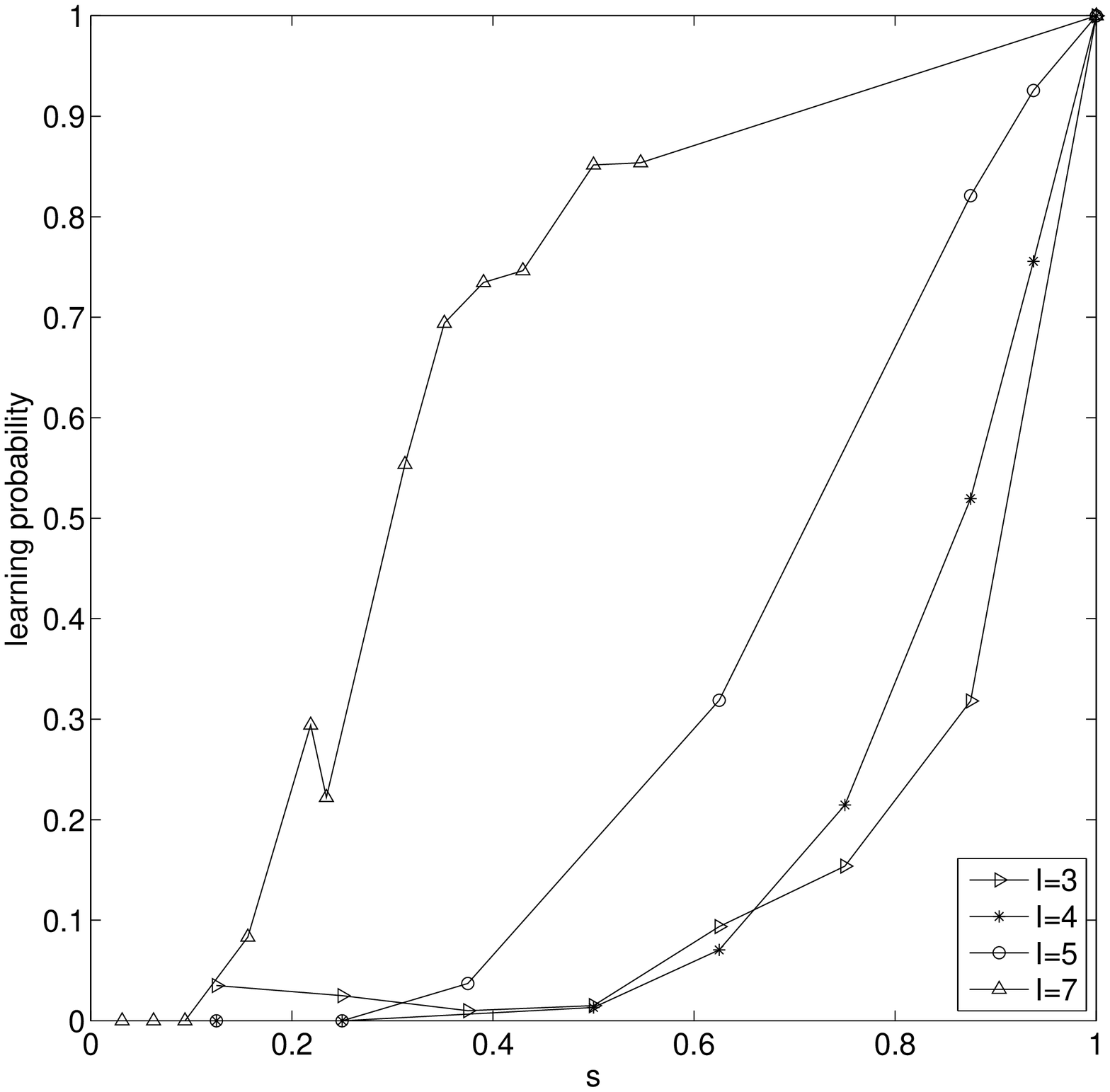}
  \includegraphics[width=.495\textwidth]{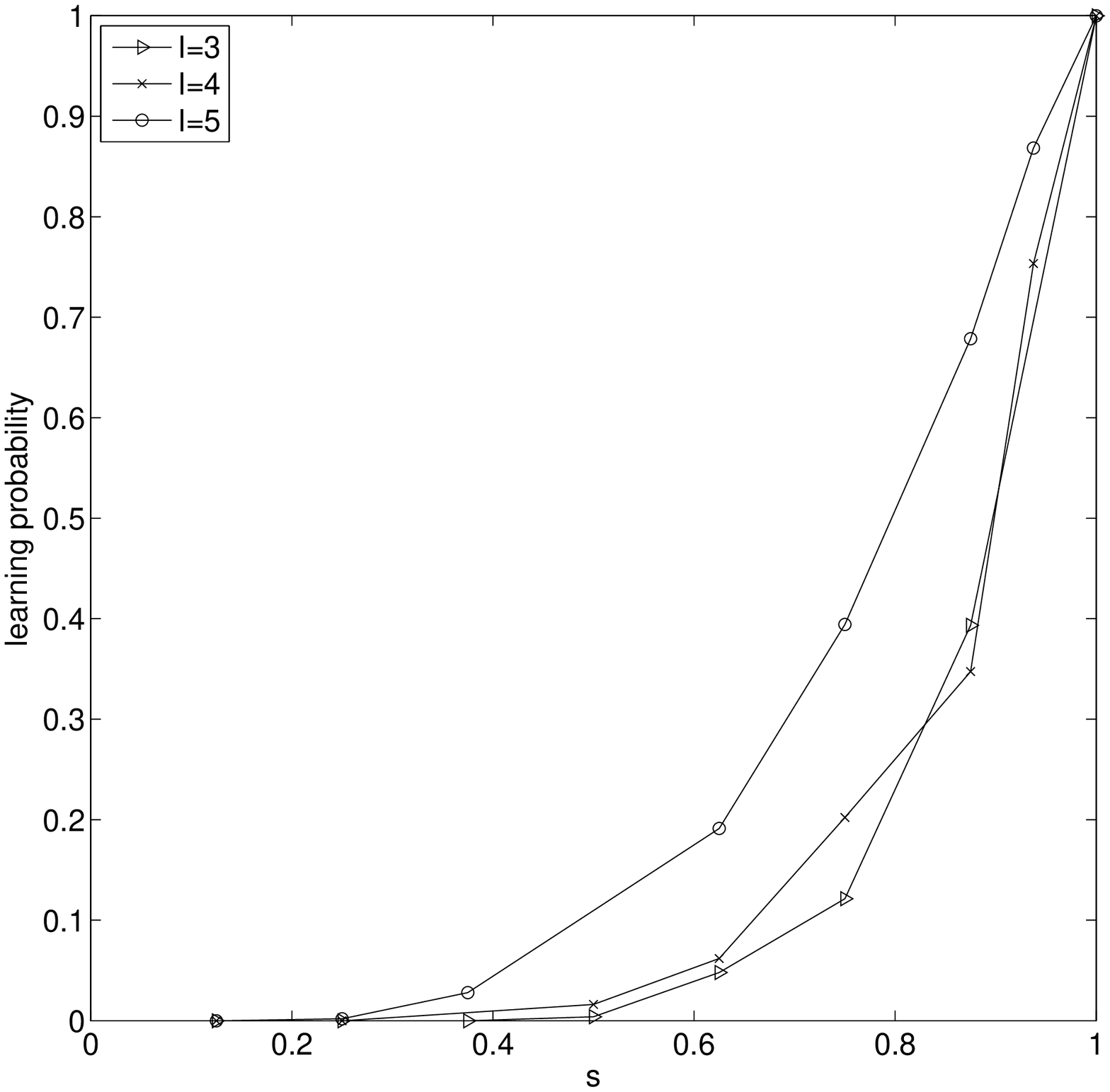}
  \caption{LEFT: The learning probability of RBNs on the even-odd task
    for different problem sizes: $I=3,4,5,7$. With increasing $I$, the
    training process requires a smaller fraction of input space in
    order to reach a higher learning probability. $N=20$,
    {$K=2.0$}, $G_{max} = 500$, init. population $=
    50$, crossover rate $= 0.7$, mutation rate $= 0.0$. We calculate
    the data over $400$ runs for all $I$s. RIGHT: The learning
    probability of RBNs on the mapping task for $I=3,4,5$. We observe
    the same behavior, but the networks generalize even worse because
    the task is more difficult. $N=40$, same GA parameters.}
  \label{fig:evenodd_1}
 \end{figure}

Next, we trained recurrent RBNs for the even-odd and the mapping
tasks.  Figure \ref{fig:evenodd_1} (LEFT) shows the learning
probability of the {$N=20$ and $K=2.0$} networks on
the even-odd task with different input sizes $I$. While the problem
size increases exponentially with $I$, we observe that despite this
state-space explosion, a higher number of inputs $I$ requires a
smaller fraction of the input space for training the networks to
achieve a high learning probability. Figure \ref{fig:evenodd_1}
(RIGHT) shows the same behavior for the mapping task, however, since
the task is more difficult, we observe a worse generalization
behavior. Also, compared to Figure \ref{fig:fig7and8}, we observe in
both cases that the generalization for recurrent networks is not as
good as for feedforward Boolean networks. In fact, for the studied
input sizes, none of the networks reaches a learning probability of
$1$ without training it on all the patterns. The lower learning
probability in RBNs is mainly due to the larger search space and the
recurrent connections, which lead to long transients and bistable
outputs that need to be interpreted in a particular way. Nevertheless,
studying adaptation and learning in RBNs, i.e., with no constraints on
the network connectivity, keeps our approach as generic as possible.

 To investigate the effect of the average connectivity $K$ on the
 learning probability, we repeat the even-odd task for networks with
 $K \in \{1.0, 1.5, 2.0, 2.5, 3.0\}$. The network size was held
 constant at $N=20$. In order to describe the training performance, we
 defined the {\em perfect training likelihood} measure $\alpha(s)$ as
 the probability for the algorithm to be able to train the network
 with the given {fraction $(s)$ of the input space (see section
   \ref{sec:definitions} for definition).

   Considering the perfect training likelihood, the results in Figure
   \ref{fig:evenodd_2} (RIGHT) show that for networks with subcritical
   connectivity $K < 2$, the patterns are harder to learn than with
   supercritical connectivity $K>2$. Close to the ``edge of chaos'',
   i.e., for $K = 2$ and $K = 2.5$, we see an interesting behavior:
   for sample sizes above $40\%$ of the patterns, the perfect training
   likelihood increases again. This transition may be related to the
   changes in information capacity of the network at $K=2$ and needs
   further investigation with different tasks.

The significant difference between the learning probability and the
perfect training likelihood for $s<0.5$ in Figure \ref{fig:evenodd_2}
is due to the small sample size. It is thus very easy for the network
to solve the task correctly, but over all $r$ runs of the experiment,
there is no network that can generalize successfully despite achieving
a perfect training score.  Also, according to the definitions in
Section \ref{sec:definitions}, it is not surprising that for a
fraction $s=1$ of the input space, i.e., all patterns are presented,
the learning probability and the perfect training likelihood are
different. Out of $r$ runs, the GA did not find perfect networks for
the task for all example, but if the networks solve the training
inputs perfectly, they will also generalize perfectly because in this
case, the training sample input includes all possible patterns.
 
\begin{figure}
  \includegraphics[width=.495\textwidth]{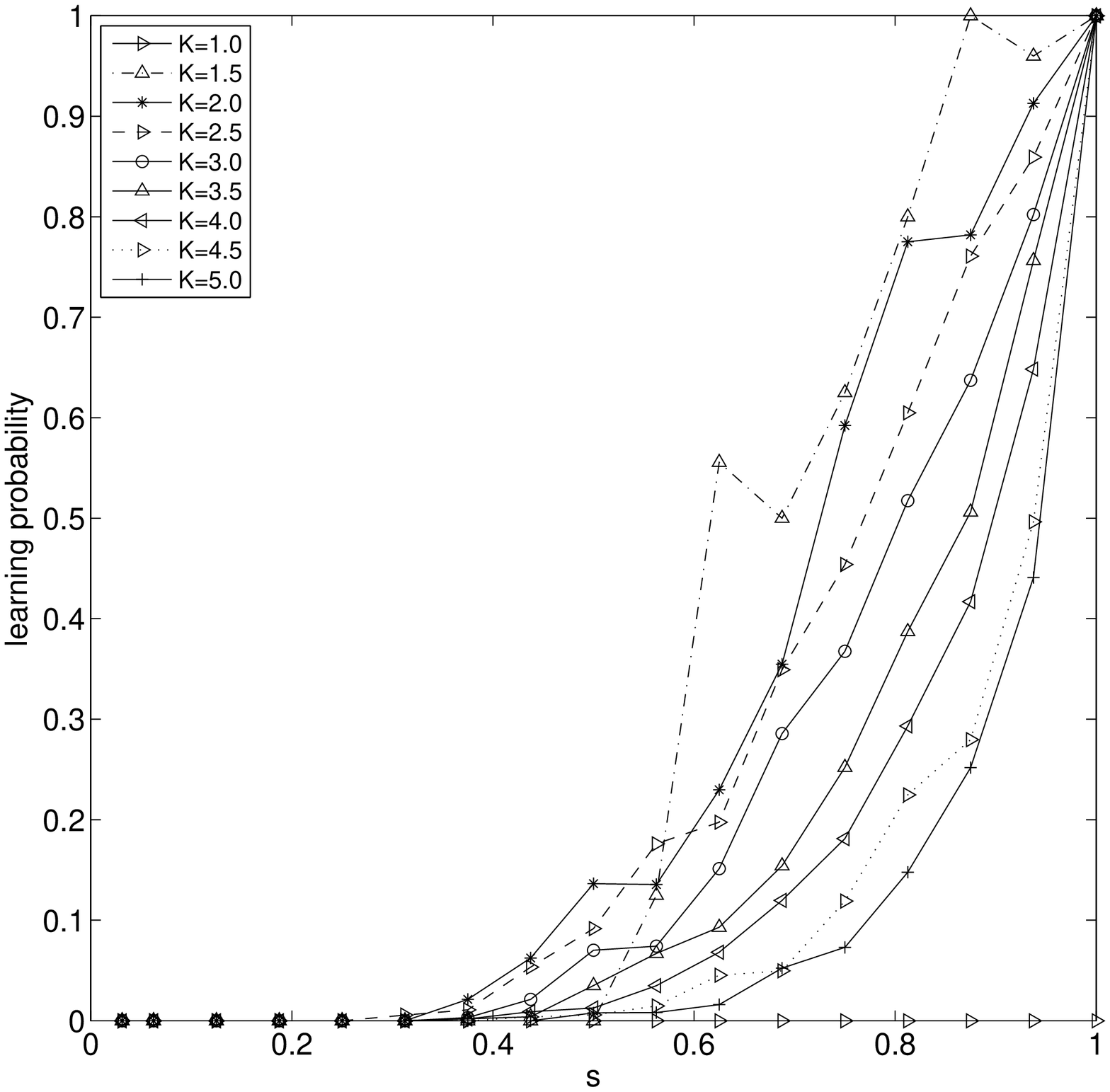}
  \includegraphics[width=.495\textwidth]{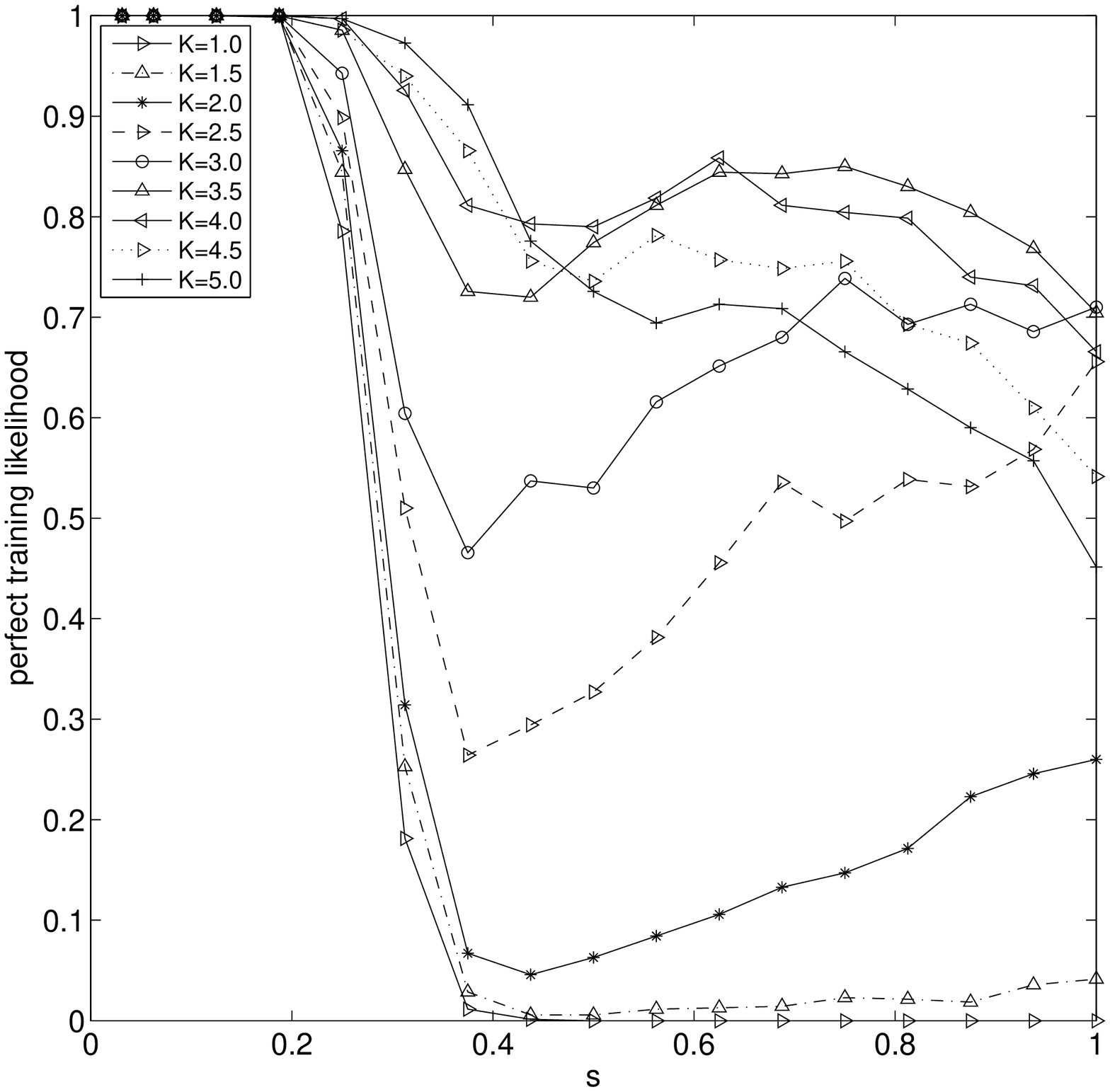}
  \caption{LEFT: The learning probability of networks with size $N=15$
    and $K \in \{1.0, 1.5, 2.0, 2.5, 3.0, 3.5, 4.5, 5.0\}$ for the
    even-odd task of size $I=5$. Networks with connectivity $K = 1.5,
    2, 2.5$ have a higher learning probability. RIGHT: The perfect
    training likelihood for the same networks and the same task. For $K \geq 2$, correctly classifying them is easier.}
   \label{fig:evenodd_2}
 \end{figure}

\section{Mean Generalization and Training Score}
\label{sec:exp4}
Figure \ref{fig:Fig14} shows the learning probability (LEFT) and the
perfect training likelihood (RIGHT) measured as Patarnello and
Carnevali did, i.e., they only counted the number of networks with
perfect generalization scores (see Section
\ref{sec:definitions}). Thus, if a network generalizes only $90\%$ of
the patterns, it is not counted in their score. That means that the
probabilistic measures of performance that we used so far have the
drawback of describing the fitness landscape of the space of possible
networks rather than the performance of a particular network, which we
are more interested in. To address this issue, we introduce a new way
of measuring both the learning and the generalization capability. We
define both of these measures as the average of the generalization and
learning fitness over $r$ runs (see section \ref{sec:definitions}).

Figure \ref{fig:Fig15} shows the generalization (LEFT) and the
training score (RIGHT) with this new measure. As opposed to Carnevali
and Patarnello's work, where higher $K$ led to a lower learning
probability, our results with the new measures for higher $K$ lead to
a higher performance with a better generalization and training score.
Our measures therefore better represent the performance of the
networks with regards to a given task because they also include
networks that can partially solve the task.

\begin{figure}
  \includegraphics[width=.495\textwidth]{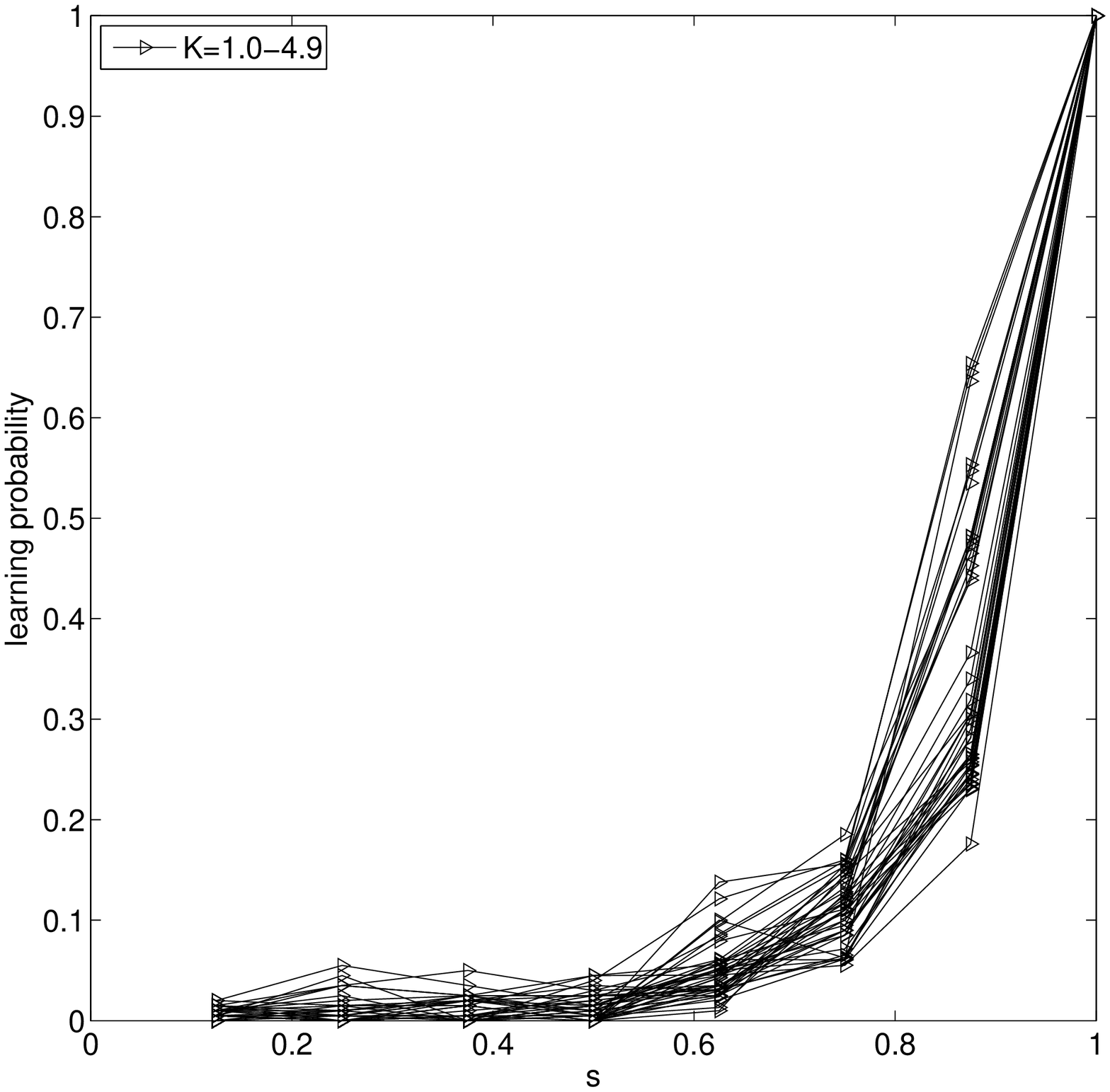}
  \includegraphics[width=.495\textwidth]{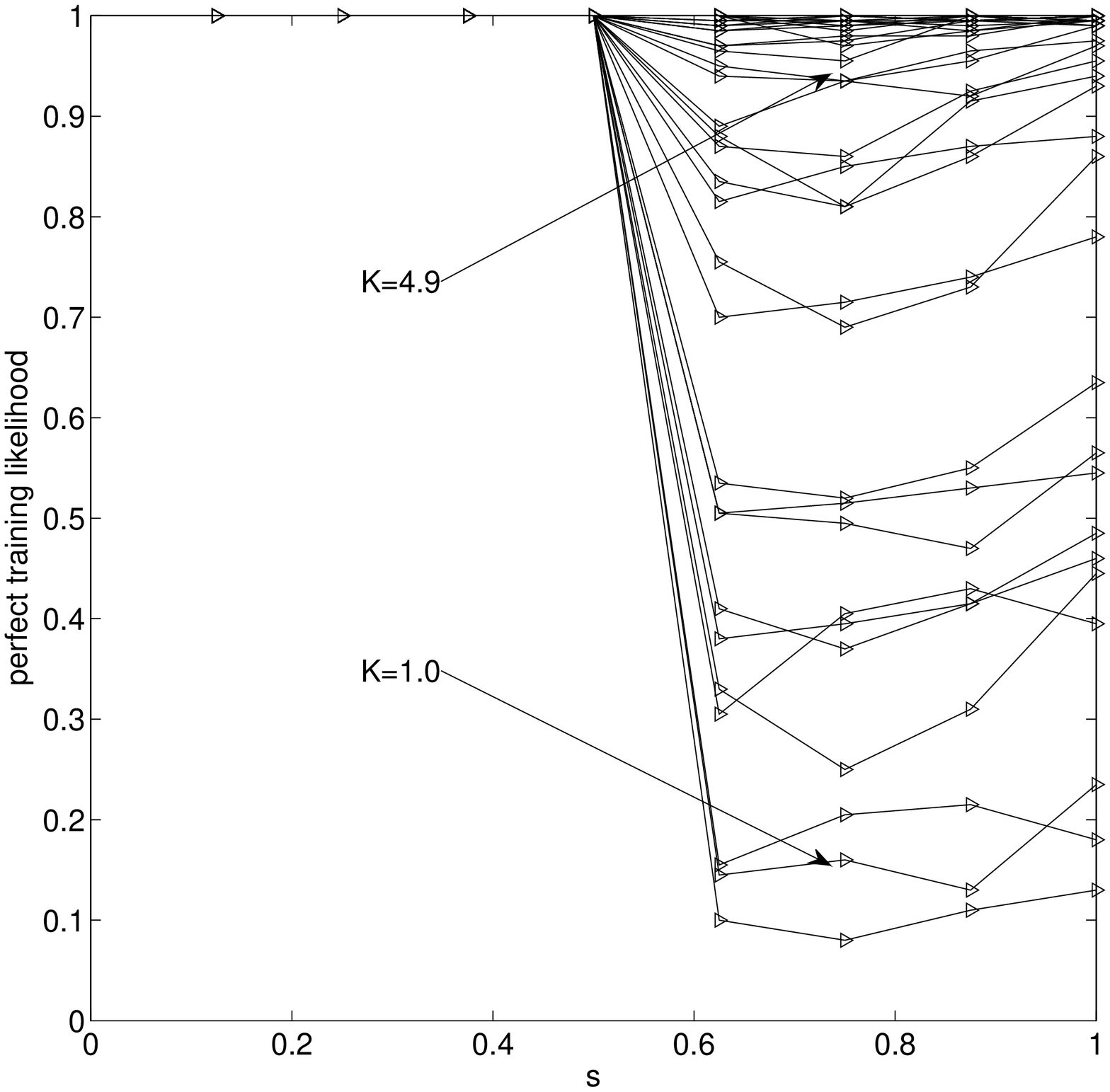}
  \caption{Learning probability (LEFT) and perfect training likelihood
    (RIGHT). $I=3, N=15$, even-odd task. Compared to the learning
    probability for $I=5$, there is not much difference between the
    learning probability of networks with various $K$ for $I=3$
    because of the small input space. However, the perfect training
    likelihood still increases with $K$. $K$ ranges from 1.0 to 4.9
    with 0.1 increments.}
  \label{fig:Fig14}
\end{figure}

\begin{figure}
  \includegraphics[width=.495\textwidth]{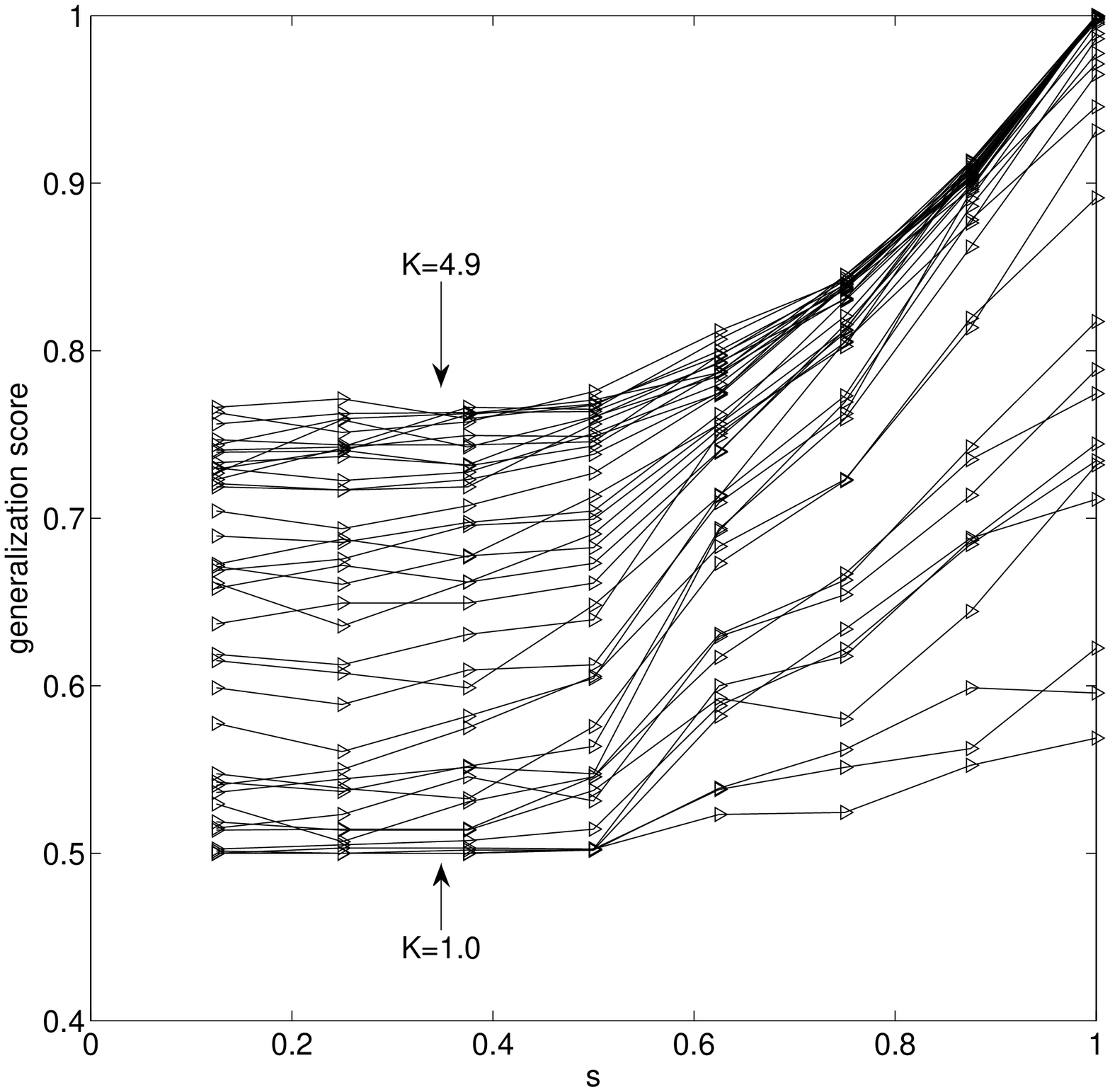}
  \includegraphics[width=.495\textwidth]{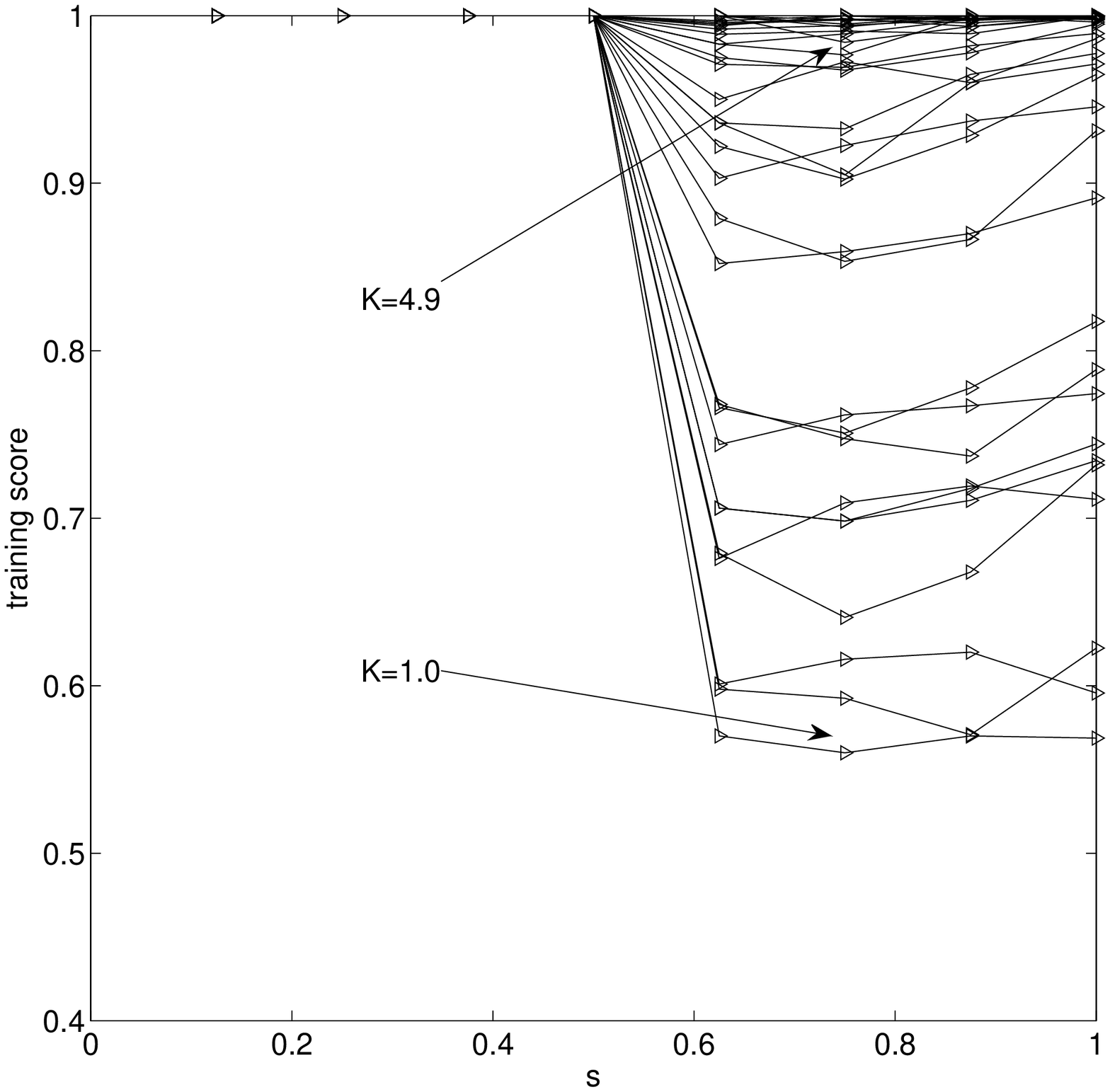}
  \caption{The new generalization (LEFT) and training score (RIGHT),
    which better reflects the performance of the networks with regards
    to a given task. $I=3, N=15$, even-odd task. $K$ ranges from 1.0
    to 4.9 with 0.1 increments.}
  \label{fig:Fig15}
\end{figure}

\section{Cumulative Measures}
\label{sec:cummeasures}
In all the previous generalization figures, the question arises which
networks are ``better'' than others, in particular if they do not
reach a maximal generalization score when less than $100\%$ of the
patterns are presented. This behavior can be observed in Figure
\ref{fig:evenodd_2} (LEFT) for the even-odd task.

Figure \ref{fig:Fig12} shows the {\em cumulative learning probability}
(LEFT) and the {\em cumulative training likelihood} (RIGHT) determined
by integrating numerically (see Section \ref{sec:definitions} for
definitions) the area under the curves of Figure
\ref{fig:Fig14}. Figure \ref{fig:Fig12} (LEFT) shows that $K$ has no
effect on the generalization and that the generalization capability is
very low. Figure \ref{fig:Fig12} (RIGHT) shows that higher $K$
increases the chance of perfect training, i.e., the network can be
trained to memorize all training patterns. Each cluster of
connectivities in Figure \ref{fig:Fig14} (RIGHT) corresponds to a
``step'' in the curves of Figure \ref{fig:Fig12} (RIGHT).

\begin{figure}
  \includegraphics[width=.495\textwidth]{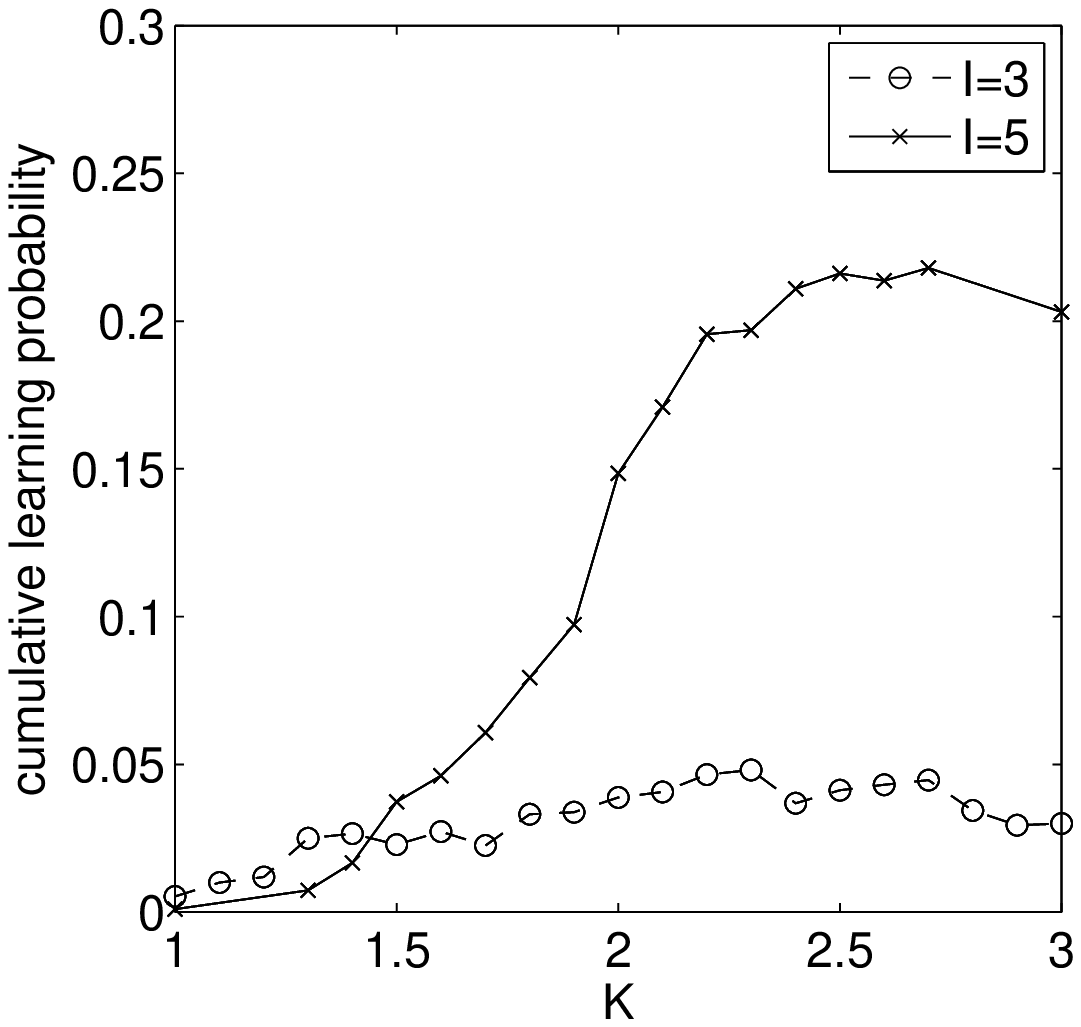}
  \includegraphics[width=.495\textwidth]{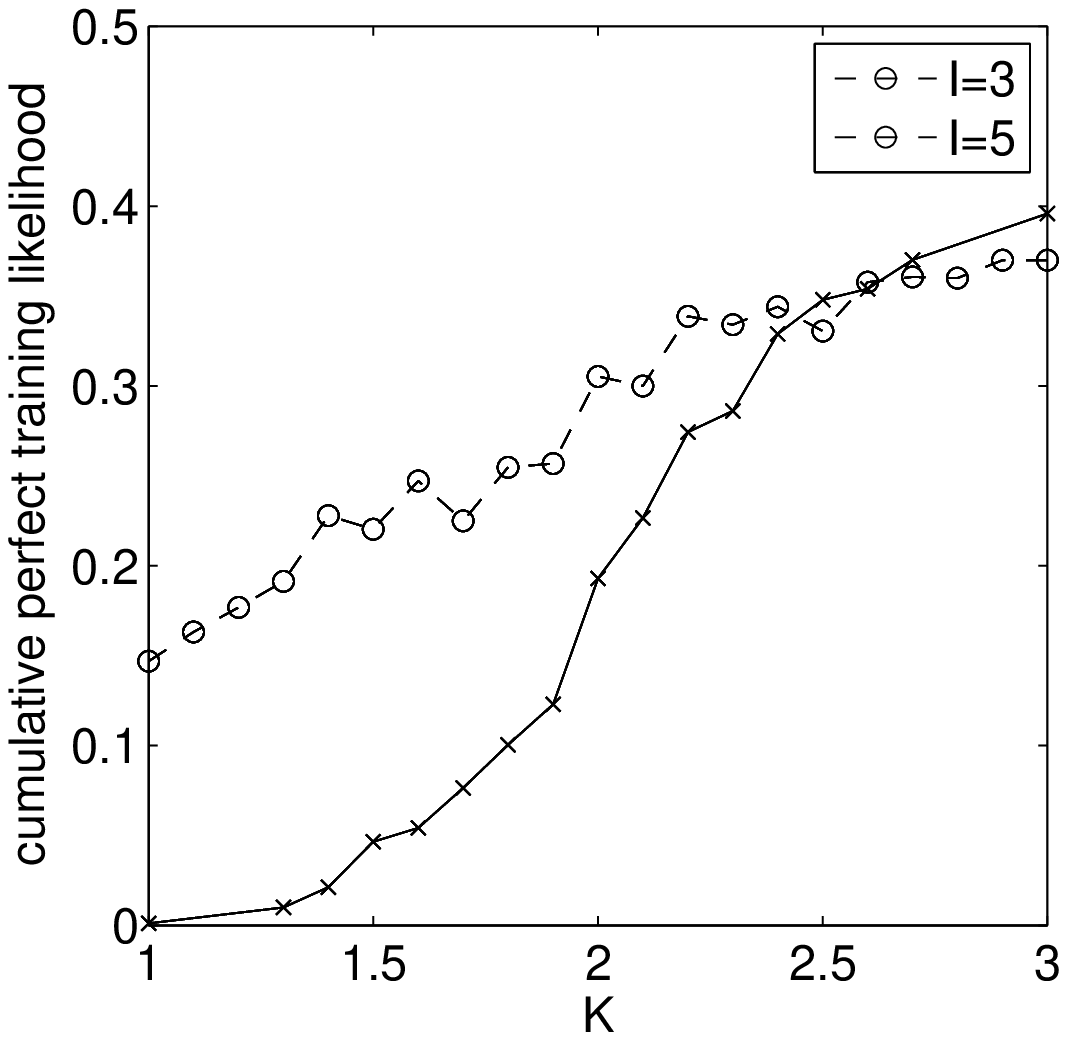}
  \caption{LEFT: Cumulative learning probability. RIGHT: Cumulative
    training likelihood. Each data point in this figure corresponds to
    the area under the curves shown in Figure \ref{fig:Fig14}. $N=15$
    in both figures, even-odd task. As one can see, perfect
    memorization is more likely with higher $K$, but perfect
    generalization is more likely for near-critical connectivity
    $1.5\leq K\leq 3$. The cumulative learning probability and the
    perfect training likelihood represent the area under the learning
    probability and perfect training likelihood curves respectively
    (see Figure \ref{fig:evenodd_2} and \ref{fig:Fig14}, and Section
    \ref{sec:definitions}).}
  \label{fig:Fig12}
\end{figure}

Figure \ref{fig:Fig13} shows the {\em cumulative generalization score}
(LEFT) and the {\em cumulative training score} (RIGHT) based on the
new measure as introduced in Section \ref{sec:exp4}. We have used the
even-odd task for two input sizes, $I=3$ and $I=5$. We observe that
$K$ has now a significant effect on the generalization score. The
higher $K$, the better the generalization. Moreover, different
intervals of $K$ result in a step-wise generalization score
increase. Figure \ref{fig:Fig13} (RIGHT) shows that the cumulative
training score for higher $K$ increases the chance of perfect
training, i.e., the network can be trained to memorize all training
patterns. Also, the higher the input size $I$, the better the
generalization, which was already observed by Patarnello and Carnevali
(see also Section \ref{sec:exp1}).

{In Section~\ref{sec:functionalentropy} we introduced
  the functional entropy as a measure of the computational richness of
  a network ensembles. {H}igher functional entropy
  implies that the probability of functions being realized by a
  network ensemble is more evenly distributed. Consequently, even if
  the target function for the training has very low probability of
  realization in an ensemble with high functional entropy, the
  evolutionary process can easily find the functions close to the
  target function. Therefore, higher functional entropy lends itself
  to higher generalization score. This fact is observable by comparing
  figures~\ref{fig:Fig13} and \ref{fig:entropyn20n100}.}

In summary, we have seen so far that according to our new measures,
higher $K$ networks both generalize and memorize better, but they
achieve perfect generalization less often. The picture is a bit more
complicated, however. Our data also shows that for networks around
$K=1.5$, there are more networks in the space of all possible networks
that can generalize perfectly. For $K>1.5$, the networks have a higher
generalization score on average, but there is a lower number of
networks with perfect generalization. That is because the fraction of
networks with perfect generalization is too small with respect to the
space of all the networks. For $K<1.5$, the networks are hard to train,
but if we manage to do so, they also generalize well.

\begin{figure}
  \includegraphics[width=.495\textwidth]{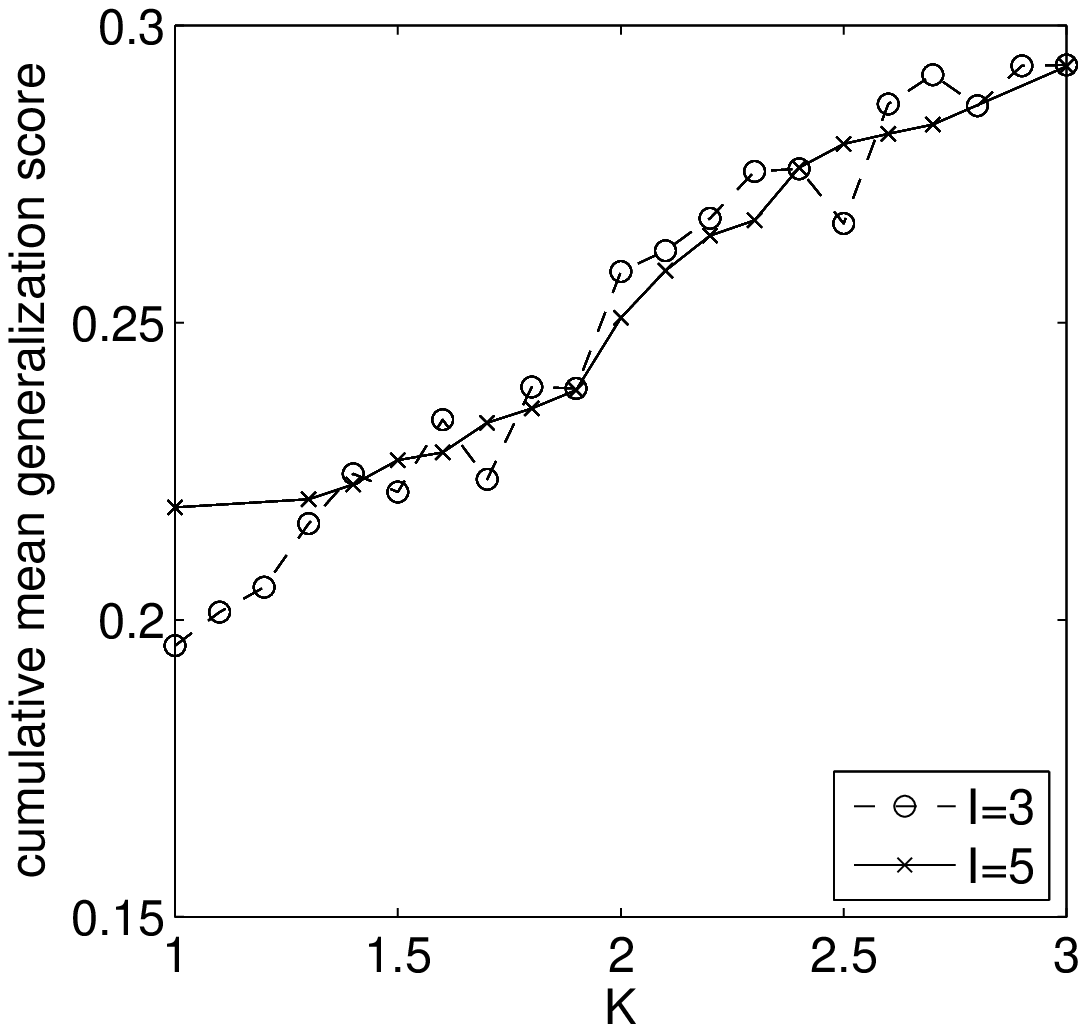}
  \includegraphics[width=.495\textwidth]{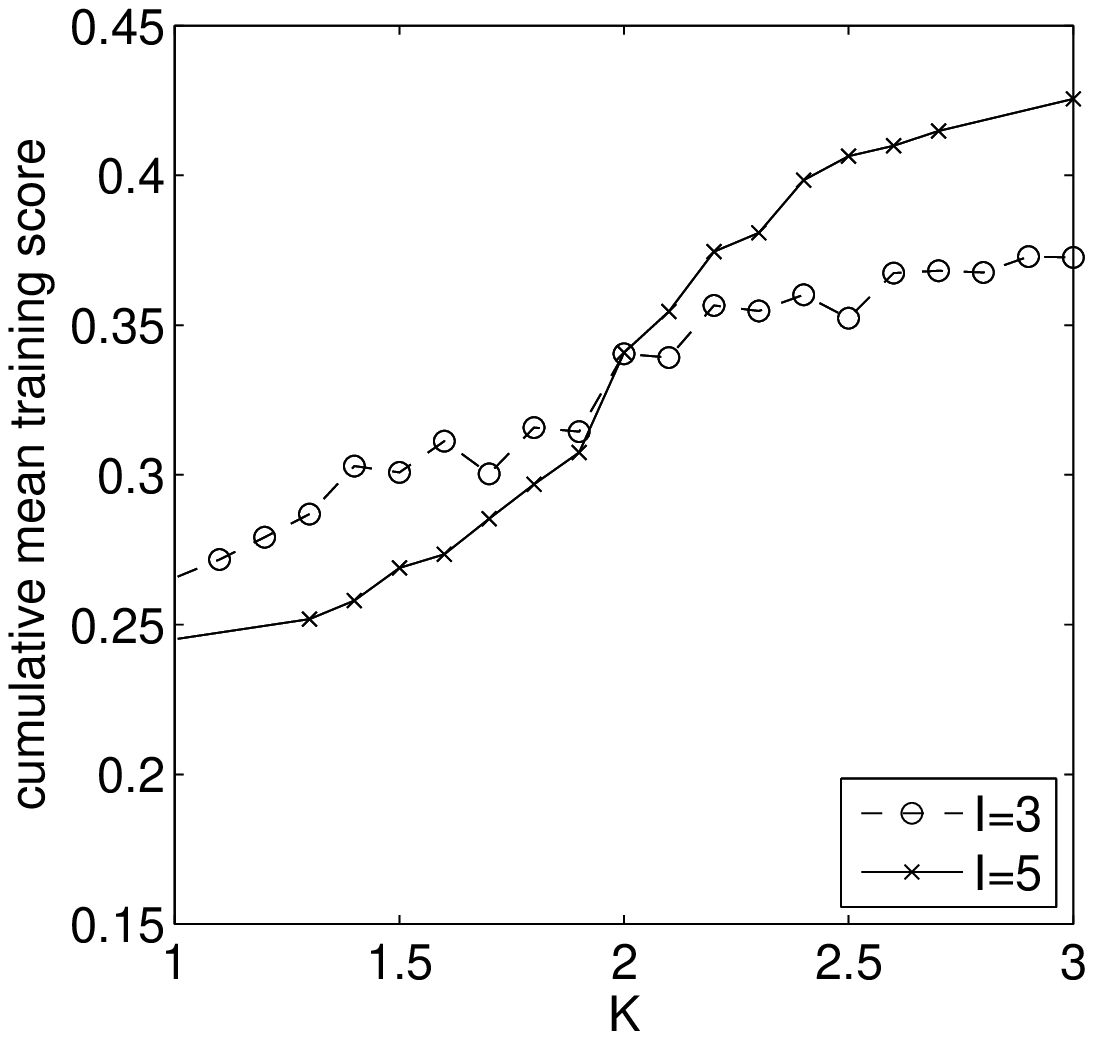}
  \caption{LEFT: Cumulative generalization score. RIGHT: Cumulative
    training score. $N=15$ in both figures, even-odd task, $I=3$ and
    $I=5$. As one can see, both the network's generalization and the
    memorization capacity increase with $K$. The cumulative
    generalization and training score represent the area under the
    mean generalization and training score curves respectively (see
    Figure \ref{fig:Fig15} and Section \ref{sec:definitions}).}
  \label{fig:Fig13}
\end{figure}

Figure \ref{fig:Fig10} shows the complete cumulative learning
probability (LEFT) and cumulative training likelihood (RIGHT)
landscapes as a function of $K$ and $N$. We observe that according to
these measures, neither the system size nor the connectivity affects
the learning probability. Also, the networks have a very low learning
probability, as seen in Figure \ref{fig:Fig12}. That means that the
performance of the training method does not depend on the system size
and the connectivity and confirms our hypothesis that Carnevali and
Patarnello's measure is more about the method than the network's
performance.

\begin{figure}
  \includegraphics[width=.495\textwidth]{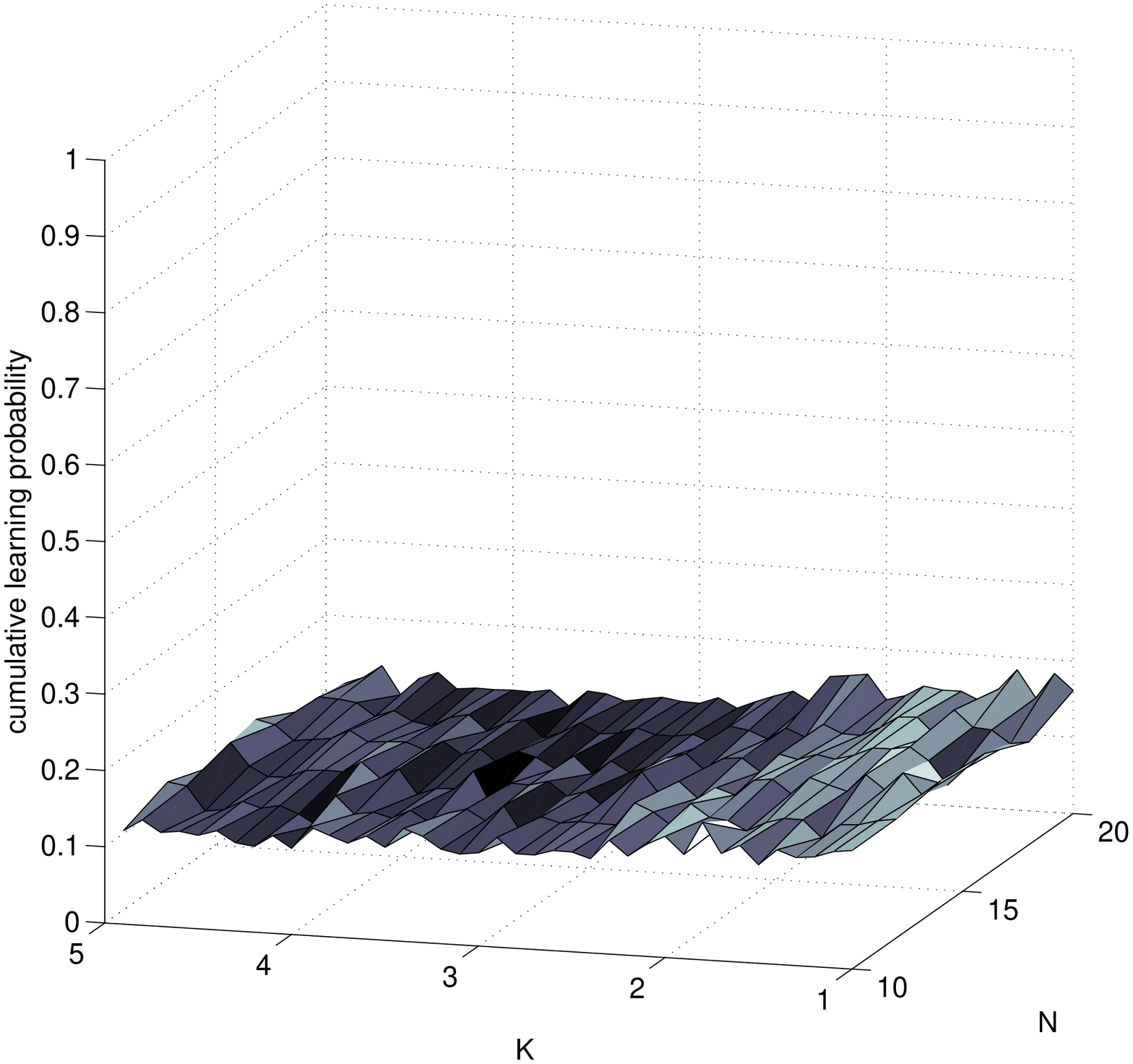}
  \includegraphics[width=.495\textwidth]{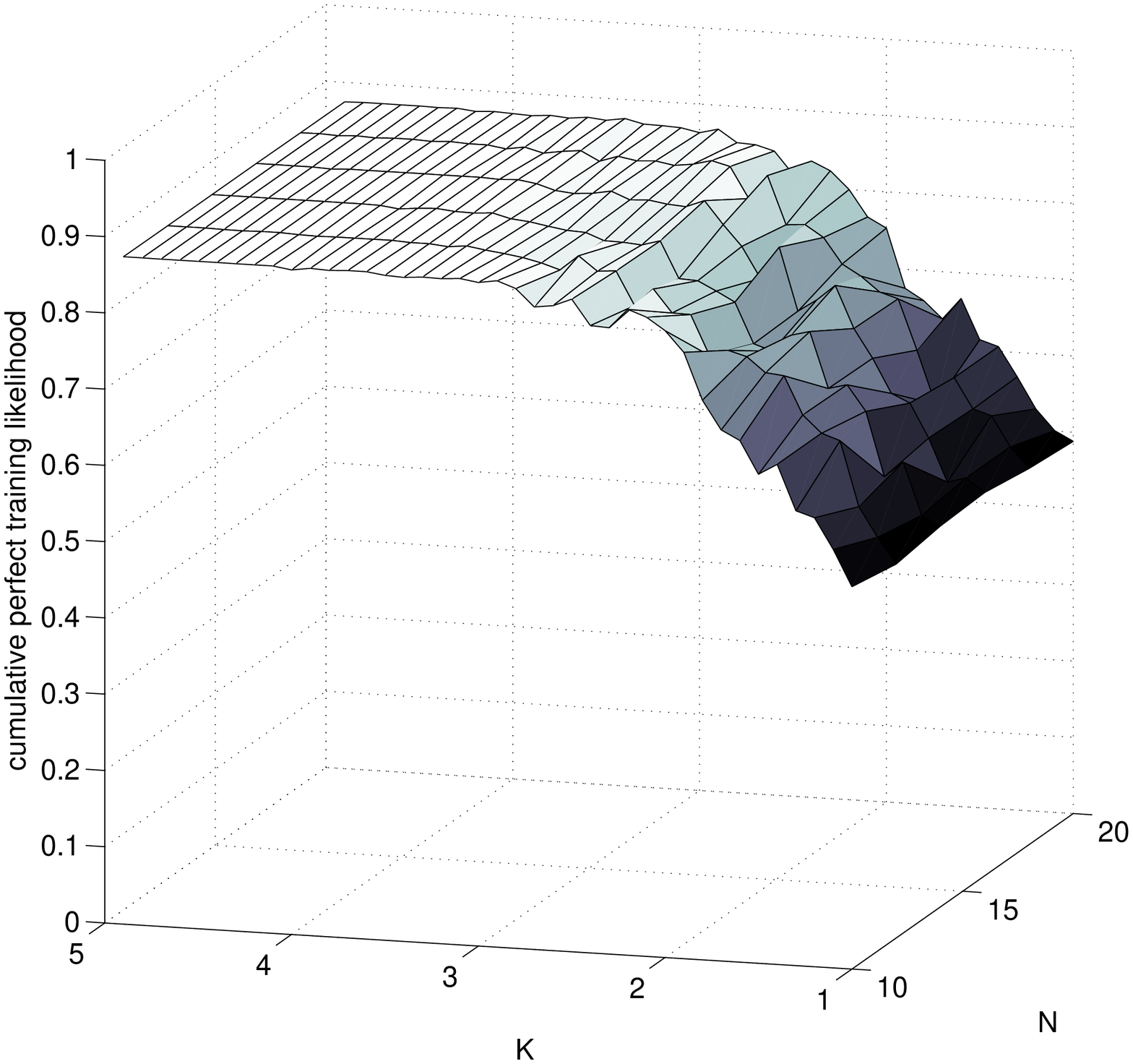}
  \caption{LEFT: Cumulative learning probability. RIGHT: Cumulative
    training likelihood. Even-odd task.}
  \label{fig:Fig10}
\end{figure}

Finally, Figure \ref{fig:Fig11} shows the same data as presented in
Figure \ref{fig:Fig10} but with our own score measures. For both the
cumulative generalization score and the cumulative training score, the
network size $N$ has no effect on the generalization and the training,
at least for this task. However, we see that for the cumulative
generalization score, the higher $K$, the higher the generalization
score. The same applies to the cumulative training score. This
contrasts what we have seen in Figure \ref{fig:Fig10}.

\begin{figure}
  \includegraphics[width=.495\textwidth]{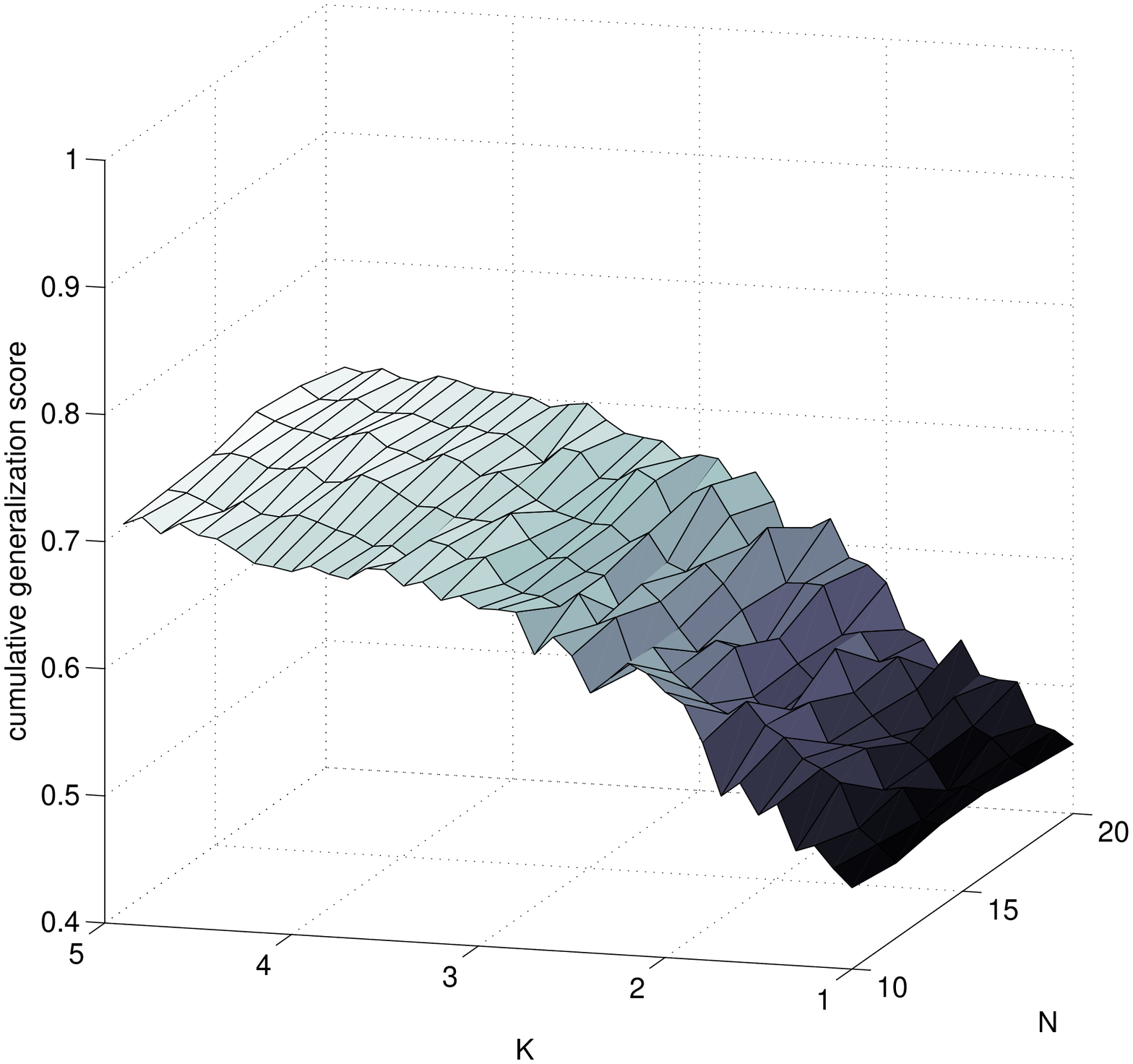}
  \includegraphics[width=.495\textwidth]{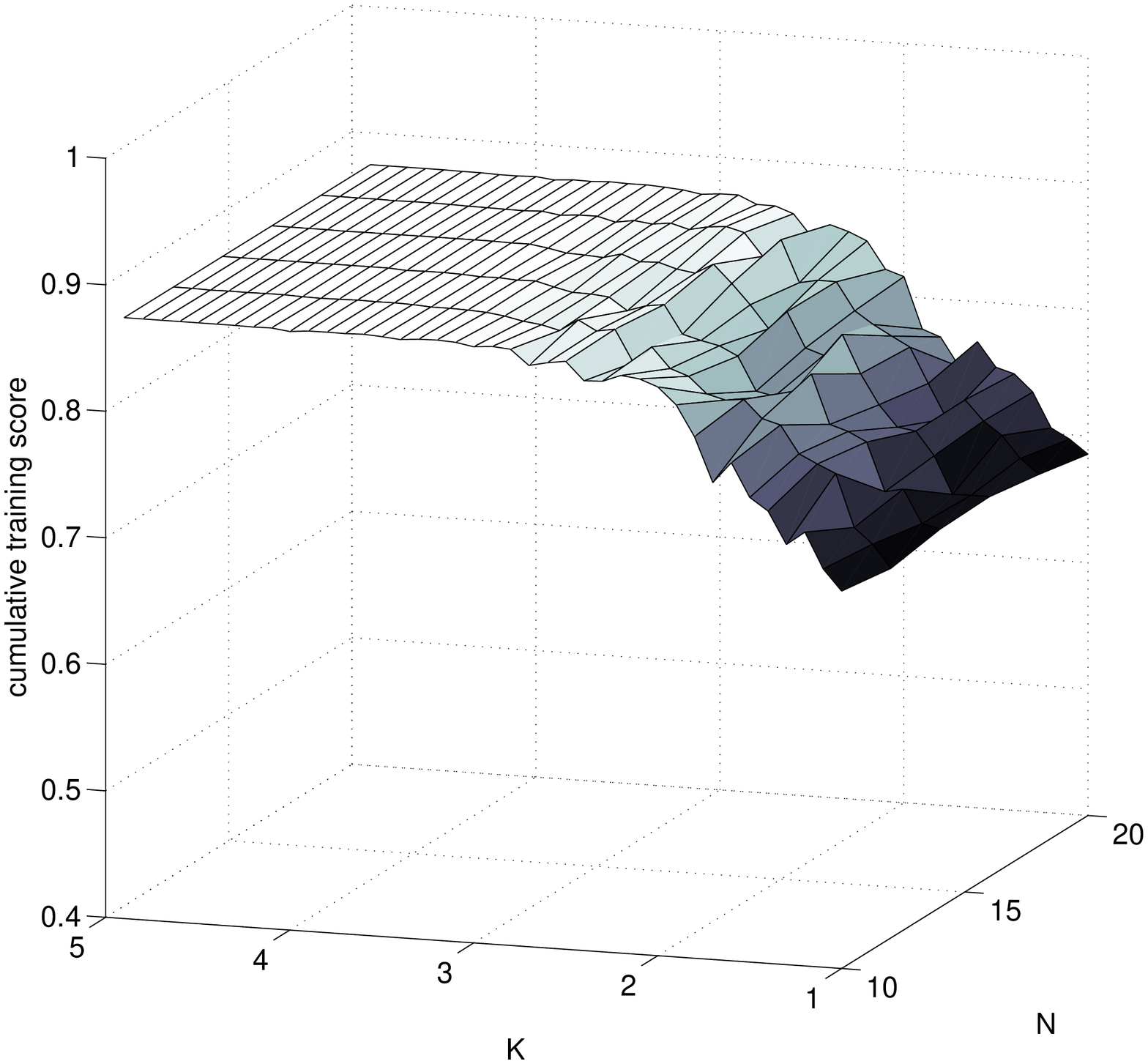}
  \caption{LEFT: Cumulative generalization score. RIGHT: Cumulative
    training score. Even-odd task.}
  \label{fig:Fig11}
 \end{figure}

\section{Discussion}
We have seen that Patarnello and Carnevali's measure quantifies the
fitness landscape of the networks rather than the network's
performance. Our newly defined measures applied to RBNs have shown
that higher $K$ networks both generalize and memorize better. However,
our results suggest that for large input spaces and for $K<1.5$ and
$K>3$ networks, the space of the possible networks changes in a way
that makes it difficult to find perfect networks (see Figures
\ref{fig:evenodd_2} and \ref{fig:Fig14}). On the other hand, for
$1.5\leq K <3$, finding the perfect networks is significantly
easier. This is a direct result of the change in the number of
possible networks and the number of networks that realize a particular
task as a function of $K$.

In \citep{lizier08_alife}, Lizier {\em et al.}  investigated
information theoretical aspects of phase transitions in RBNs and
concluded that subcritical networks ($K<2$) are more suitable for
computational tasks that require more of an information storage, while
supercritical networks ($K>2$) are more suitable for computations that
require more of an information transfer. The networks at critical
connectivity ($K=2$) showed a balance between information transfer and
information storage. This finding is purely information theoretic and
does neither consider input and outputs nor actual computational
tasks. In our case, solving the tasks depends on the stable network
states and their interpretations. The results in \citep{lizier08_alife}
do not apply directly to the performance of our networks, but we
believe there is a way to link the findings in future work.  Compared
to Lizier {\em et al.}, our experiments show that supercritical
networks do a better job at both memorizing and generalizing. However,
from the point of view of the learning probability, we also observe
that for networks with $1.5\leq K < 3$, we are more likely to find
perfect networks for our specific computational tasks. 

{We measured the computational richness of a network
  ensemble by using its functional entropy
  (Section~\ref{sec:functionalentropy}). In
  Section~\ref{sec:cummeasures}, we explained how higher functional
  entropy for a network ensemble result in higher generalization
  score. In addition to higher generalization score, higher functional
  entropy improves the performance of an evolutionary search because
  it naturally result in higher fitness diversity in the evolving
  population (Figure~\ref{fig:fitstdev}). With more evenly distributed
  probability of functions realized by the individual networks in the
  ensemble, it is more likely that individuals in the population
  realize different functions and thus diversifying the fitness of the
  population. This fitness diversity creates higher gradient that
  increase the rate of fitness improvement during the evolution
  \citep{PRICE:1972p2444}.}
\begin{figure}
\centering
  \includegraphics[width=.50\textwidth]{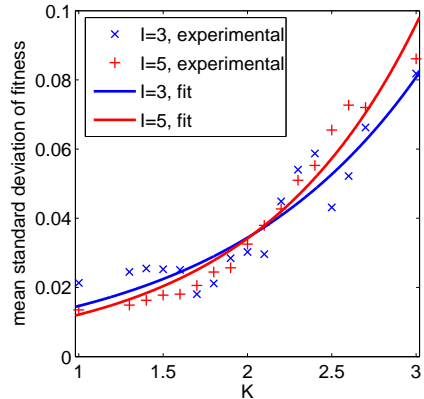}
  \caption{{Mean fitness standard deviation of the
      population for $I=3$ and $I=5$. $N=20$, $1.0\le K\le 3.0$. An
      increase in the functional entropy of the network increases the
      diversity of the fitness of the population and creates gradients to guide
      the search process (cf. figures~\ref{fig:entropyn20n100} and \ref{fig:Fig11}).}}
  \label{fig:fitstdev}
 \end{figure}
\section{Conclusion}
In this paper we empirically showed that random Boolean networks can
be evolved to solve simple computational tasks. We have investigated
the learning and generalization capabilities of such networks as a
function of the system size $N$, the average connectivity $K$, problem
size $I$, and the task. We have seen
that the learning probability measure used by
 \cite{patarnello87:europhys} was of limited use and have thus
introduced new measures, which better describe what the networks are
doing during the training and generalization phase. The results
presented in this paper are invariant of the training parameters and
are intrinsic to both the learning capability of dynamical automata
networks and the complexity of the computational task. Future work
will focus on the understanding of the Boolean function space, in
particular on the function bias.


\section*{Acknowledgments}
This work was partly funded by NSF grant \# 1028120.

\bibliographystyle{plainnat}
\bibliography{rbn}

\end{document}